\documentclass{article} % For LaTeX2e
\usepackage[final]{colm2026_conference}

\usepackage{microtype}
\usepackage{hyperref}
\usepackage{url}

\usepackage{graphicx}
\usepackage{tabularx}
\usepackage[table]{xcolor} % loads colortbl internally on many setups; if not, add 
\usepackage{makecell}
\usepackage{enumitem}

\usepackage{xspace}  % 
\newcommand{\webarena}{\textsc{WebArena}\xspace}
\newcommand{\visualwebarena}{\textsc{VisualWebArena}\xspace}

\usepackage{booktabs}
\usepackage{multirow}
\usepackage{array}
\usepackage{colortbl}
\usepackage{wrapfig}
\usepackage{pifont}

\usepackage{amsmath}
\usepackage{bm}
\usepackage[most]{tcolorbox}

\usepackage{cleveref}  
\crefname{section}{§}{§§}
\definecolor{ForestGreen}{HTML}{009B55}
\definecolor{OrangeRed}{HTML}{c22f2f}
\definecolor{ToolBlockGray}{HTML}{EEEEEE} % slightly darker than F2F2F2 to avoid "white seams"

\newcommand{\MyUpArrow}[1]{\textcolor{ForestGreen}{$\uparrow$\,#1}}
\newcommand{\MydownArrow}[1]{\textcolor{OrangeRed}{$\downarrow$\,#1}}

\newcommand{\GroupHdrFont}{\large\bfseries}
\newcommand{\ColHdrFont}{\small\bfseries}
\newcommand{\AVGHdrFont}{\large\bfseries}
\newcommand{\AVGNumFont}{\large}  % \normalsize
\newcommand{\DeltaFont}{\footnotesize}

\newcolumntype{L}[1]{>{\raggedright\arraybackslash}m{#1}}
\newcolumntype{C}[1]{>{\centering\arraybackslash}m{#1}}
\newcolumntype{R}[1]{>{\raggedleft\arraybackslash}m{#1}}

\newcommand{\ToolBlockLabelH}[2]{%
  \makebox[2.6em][c]{%
    \rotatebox[origin=c]{90}{\parbox[c][#2][c]{#2}{\centering\normalsize #1}}%
  }%
}

\newcommand{\AVGCellPlain}[1]{\AVGNumFont #1}
\newcommand{\AVGCellDelta}[2]{\makecell[r]{\AVGNumFont #1\\{\DeltaFont (#2)}}}

\definecolor{AVGGray}{gray}{0.93}   %
\definecolor{LightGrayRow}{gray}{0.93}

\usepackage{listings}
\newcommand{\walt}{\texttt{WALT}}
\newcommand{\skillweaver}{\texttt{SkillWeaver}}
\newcommand{\hybrid}{\texttt{Hybrid-Agent}}
\newcommand{\webmcp}{\texttt{WebMCP}}

\usepackage{xcolor}

\definecolor{WALTColor}{RGB}{255,230,200} % 
\definecolor{SkillWeaverColor}{RGB}{230,245,255}   % 
\definecolor{HybridColor}{RGB}{230,255,230}        % 
\definecolor{WebMCPColor}{RGB}{255,235,230}        % 
\definecolor{ShallowBlue}{HTML}{ddf2ff}
\definecolor{ShallowPink}{HTML}{fadfdc}
\definecolor{ShallowGreen}{HTML}{cff0da}
\definecolor{ShallowOrange}{HTML}{fff1b9}

\newcommand{\waltcolor}{
\colorbox{ShallowPink}{\raisebox{0pt}[0.7\height]{\texttt{WALT}}}
}
\newcommand{\skillweavercolor}{
\colorbox{ShallowGreen}{\raisebox{0pt}[0.7\height]{\texttt{SkillWeaver}}}
}
\newcommand{\hybridcolor}{
\colorbox{ShallowBlue}{\raisebox{0pt}[0.7\height]{\texttt{Hybrid-Agent}}}
}
\newcommand{\webmcpcolor}{
\colorbox{ShallowOrange}{\raisebox{0pt}[0.7\height]{\texttt{WebMCP}}}
}

\usepackage{lineno}

\definecolor{darkblue}{rgb}{0, 0, 0.5}
\hypersetup{colorlinks=true, citecolor=darkblue, linkcolor=darkblue, urlcolor=darkblue}

\newtcolorbox{promptbox}{
  colback=white,
  colframe=black!50,
  boxrule=0.4pt,
  arc=3pt,
  left=6pt,
  right=6pt,
  top=6pt,
  bottom=6pt,
  enhanced,
  breakable
}

\lstdefinestyle{promptstyle}{
  basicstyle=\ttfamily\small,
  breaklines=true,
  breakatwhitespace=false,
  columns=fullflexible,
  keepspaces=true,
  showstringspaces=false,
  frame=none
}

\usepackage[T1]{fontenc}
\usepackage[utf8]{inputenc}

\definecolor{TakeawayBG}{HTML}{EDF4FF}
\definecolor{TakeawayBorder}{HTML}{8FB1DC}
\definecolor{TakeawayBar}{HTML}{CFE0F7}
\definecolor{TakeawayTitleText}{HTML}{173A63}
\definecolor{TakeawayHeadText}{HTML}{204A87}

\newtcolorbox{takeawaycard}[1][]{
  enhanced,
  breakable,
  colback=TakeawayBG,
  colframe=TakeawayBorder,
  boxrule=0.8pt,
  arc=2pt,
  left=8pt,
  right=8pt,
  top=8pt,
  bottom=8pt,
  #1
}
\newcommand{\takeawayhead}[1]{%
  \vspace{4pt}%
  {\noindent\color{TakeawayHeadText}\bfseries\itshape #1\par}%
  \vspace{1pt}%
}

\newlist{takeawayitems}{itemize}{1}
\setlist[takeawayitems]{
  label=\textbullet,
  leftmargin=1.6em,
  itemsep=2pt,
  topsep=2pt,
  parsep=0pt,
  partopsep=0pt
}

\title{The Tool Illusion: Rethinking Tool Use in Web Agents}
\author{Renze Lou\textsuperscript{\rm $\ddagger$}\thanks{This work was done during Renze's internship at Microsoft Research.} \quad
Baolin Peng\textsuperscript{\rm $\dagger$} \quad
Wenlin Yao\textsuperscript{\rm $\dagger$} \quad
Qianhui Wu\textsuperscript{\rm $\dagger$} 
\\
\textbf{Hao Cheng}\textsuperscript{\rm $\dagger$} \quad
\textbf{Suman Nath}\textsuperscript{\rm $\dagger$} \quad
\textbf{Wenpeng Yin}\textsuperscript{\rm $\ddagger$} \quad
\textbf{Jianfeng Gao}\textsuperscript{\rm $\dagger$}
\\ 
\\
\textsuperscript{\rm $\dagger$}Microsoft Research, Redmond
  \\
  \textsuperscript{\rm $\ddagger$}The Pennsylvania State University, University Park
  \\
  {\small \texttt{renze.lou@psu.edu, baolinpeng@microsoft.com}}
}

\begin{document}

\ifcolmsubmission
\linenumbers
\fi

\maketitle

\begin{abstract}
As web agents rapidly evolve, an increasing body of work has moved beyond conventional atomic browser interactions and explored tool use as a higher-level action paradigm. 
Although prior studies have shown the promise of tools, their conclusions are often drawn from limited experimental scales and sometimes non-comparable settings. As a result, several fundamental questions remain unclear: i) whether tools provide consistent gains for web agents, ii) what practical design principles characterize effective tools, and iii) what side effects tool use may introduce. To establish a stronger empirical foundation for future research, we revisit tool use in web agents through an extensive and carefully controlled study across diverse tool sources, backbone models, tool-use frameworks, and evaluation benchmarks. 
Our findings both revise some prior conclusions and complement others with broader evidence.
% Our results both revise some conclusions suggested by prior work and complement others with broader evidence.
We hope this study provides a more reliable empirical basis and inspires future research on tool-use web agents.
% useful insights for 
\end{abstract}

\section{Introduction}
\label{sec:intro}

Large language model agents have recently shown strong potential in real-world interactive environments, such as computer-use or browser-use settings~\citep{xie2024travelplanner,xie2024osworld,deng2023mind2web,yang2025ultracua,zhang2025agent}. In the web domain, early work typically relies on low-level atomic actions that mimic human interaction with webpages, such as mouse \texttt{click} and keyboard \texttt{type}~\citep{yao2022webshop,zhouwebarena}. However, this human-like action paradigm is often brittle and inefficient for agent: low-level browser actions are error-prone, and each action requires separate reasoning and decision making, substantially increasing the agent’s web exploration burden~\citep{koh2024tree,sodhistep}.
% At the same time, exploring the web through such low-level actions can be slow and error-prone, making it harder for the agent to efficiently discover successful trajectories for completing tasks.
% 
% for web agent, as the web behaviour can be dynamically and complex, while it burden the reaonsing effort for agent  

To address this limitation, recent works have increasingly adopted tools, typically application programming interfaces (APIs) or procedural functions, as higher-level action abstractions~\citep{krishna2024paffa,wang2025inducing}.
These tools either expose direct interfaces to the backend of web services (developed by humans), or encapsulate reusable sequences of low-level browser operations (synthesized by agent).
% These tools either expose direct interfaces to the backend of web services (developed by human)~\citep{song2025beyond,webmcp2025} or encapsulate reusable sequences of low-level browser operations that is synthezid by agent itself~\citep{zheng2025skillweaver,prabhu2025walt}. 
Consequently, a single tool call can often substitute for a long sequence of atomic webpage interactions, and prior work has demonstrated the promise of this paradigm in improving web agent performance~\citep{song2025beyond,zheng2025skillweaver,prabhu2025walt}.

However, our understanding of tool use in web agents, especially LLM-synthesized tools, remains incomplete. Prior work is often conducted at limited experimental scale, typically using only a single tool source (e.g., tools synthesized by one model) and evaluating it on only a small set of backbone models (from the same model family). As a result, many existing conclusions are based on relatively narrow evidence and do not fully reveal how tool utility varies across tool sources, backbone models, and evaluation settings. Moreover, this limitation renders several prior findings incomplete, subjective, or even mutually inconsistent. For example, \citet{zheng2025skillweaver} synthesize tools solely with GPT-4o and evaluate them only on homogeneous backbone models (GPT-4o and GPT-4o-mini), concluding that weaker models benefit more from tools, whereas \citet{prabhu2025walt}, using a different tool source and a different set of backbone models, report the opposite trend. Such inconsistencies leave future research on tool-use web agents without a sufficiently solid foundation and make progress in this area potentially misleading.
% making progress in this area both difficult

% indicate that the current literature still lacks a large-scale, systematic, and apples-to-apples study of tool use in web agents.
% It is still unclear when tools help, how much they help, and what side effects they may introduce. 

 % Although all of them are designed for tool-use web automation, they differ substantially in their technical design. For example, while both \skillweaver~and~\walt~build their tools on top of Playwright,\footnote{\url{https://playwright.dev/}}
 % they differ considerably in tool distribution and format design. This diversity makes it challenging for future developers to build effective tool-use web agents, as existing research has not yet yielded clear and concrete design conclusions.\footnote{\cite{prabhu2025walt} reports performance comparisons across different frameworks; however, those results are not based on a strictly apples-to-apples evaluation, which makes some of the resulting conclusions ambiguous and potentially debatable. See Appendix~\ref{appendix:frameowkr_details_tool_exp} for further discussion.}

To offer a reliable empirical basis for future research, we revisit tool use in web agents.
% a solid empirical foundation and 
% through comprehensive and carefully controlled experiments
Particularly, we ask three high-level questions: i) \textit{Do tools, especially LLM-synthesized ones, consistently improve over browser-use agents, and under what conditions do they help or fail?} ii) \textit{What practical design principles characterize effective tools?} iii) \textit{What costs and side effects do tools introduce compared with low-level browser interaction?}
% Rather than taking prior claims at face value, we systematically examine tool use under a fair and consistent evaluation protocol.
We therefore conduct an extensive empirical study across multiple dimensions, including diverse tool sources, five backbone models, three tool‑use agent frameworks, and two realistic web benchmarks, supplemented with a series of controlled analyses. Notably, some of our key observations differ from (and in some cases revise) prior conclusions, while others complement and extend them.
We summarize the key findings of this work below to aid future research:
% \begin{itemize}[leftmargin=2em,label=\ding{118}]
% \itemsep0.2em
% \item XXX
% \item YYY
% \item ZZZ
% \end{itemize}

\begin{itemize}[leftmargin=2em,label=\ding{118},itemsep=0pt,topsep=2pt,parsep=0pt,partopsep=0pt]
    \item \textbf{LLM-synthesized tools mainly exhibit one-way capability distillation.} They provide consistent gains primarily when the tool user is clearly weaker than the tool synthesizer; for stronger models, the benefits are limited, inconsistent, and often negative.

    \item \textbf{More comprehensive tools are not necessarily better.} Overly task-specific tools often hurt generalizability and utility, and some complex tools may rarely or never be invoked in practice. A better design principle is to reserve tools for deterministic UI operations while leaving grounding-dependent reasoning and planning to the browser agent.
    % Overly task-specific, end-to-end tools often harm generalizability and utility--some complex tools even never invoked by agent. A better design principle is to encapsulate only deterministic UI operations as tools, while leaving grounding-dependent reasoning and planning to the browser agent.
    \item \textbf{Effective tool design should prioritize functional coverage and compositionality.} Less comprehensive tools can still be highly effective, as long as they compose well and collectively cover frequent real-world user intentions.

    \item \textbf{Tool entails hidden tax.} Although tools can replace long sequences of browser actions, large tool libraries and low-utility synthesized tools may increase both token cost and action overhead.

    \item \textbf{Semantic skills offer a flexible alternative to programmatic tools.} Their transparency can mitigate some drawbacks of black-box tools, provided that the backbone model is strong enough to interpret and use them effectively.

    \item \textbf{Vision remains useful in browser- and tool-use settings.} Visual grounding continues to improve performance, while tool use makes agents more robust to its absence.
\end{itemize}

\section{Related Work}
\label{sec:related_work}

% \vspace{-0.8em}

\noindent\textbf{Browser-Use Web Agents.}
% Large Language Models (LLMs) web agents are emerged recently for facilitating 
Web agents, which use large language models (LLMs) or vision-language models (VLMs) to operate web browsers and automate user tasks, have received increasing attention in recent years~\citep{cheng2024seeclick,gu2024your,zheng2024gpt,he2024webvoyager,he2025openwebvoyager}. Given webpage observations such as the accessibility tree, DOM, and screenshots, these agents execute human-like atomic actions (e.g., \texttt{click}, \texttt{type}) to interact with the web~\citep{yao2022webshop,deng2023mind2web,wu2025webwalker}. Existing browser-use web agents can be broadly grouped into two directions. a) \textbf{Prompting-based agents with enhanced planning and search.} For example, \cite{koh2024tree} and \cite{putta2024agent} use tree‑search–based methods (e.g., Monte Carlo Tree Search) to enhance test‑time exploration of web agents; \cite{sodhistep}~propose policy stacks for handling complex web behaviors beyond direct prompting; \cite{yang2023set} leverages Set-of-Mark (SoM) for additional visual grounding. b) \textbf{Learning web tasks via fine-tuning}. Another line of work improves web agents through domain-specific training, for example by instruction-tuning multimodal foundation models~\citep{furutamultimodal} or synthesizing high-quality trajectories for web-agent tuning~\citep{pahuja2025explorer,wang2025adapting,hongjinlearn,liu2025webexplorer}.

% qiu2025alitaG,qiu2025alita

% Web browser agent, which set large language model (LLM)/Vision Lanaguge Models (VLM) agent for the user browser and facilitating web task automation, has drawed extensive attentions recently. By using the web page accessibility tree (axtree), HTML content (DOM), and optional page screenshot, browser agent aopt human-like atomic actions (e.g., \texttt{click}, \texttt{type}) to help user interact with the web. Existing web agent systems can be devided into two main streamlines: a) Use various prompting strategies with advanced planning and search quippped with the agent. For example, \cite{koh2024tree} and \cite{putta2024agent} use tree search and Monte Carlo Tree Search to enhance the test-time exploration for web agent; \cite{sodhistep} propose policy stack to address the complex web behaviors that are hard to addressed by direct prompting; \cite{yang2023set} leverage Set-of-Mark (SoM) as additional visual grounding. b) Fine tuning for stronger backbones for web agent. For instance, \cite{furutamultimodal} train the multi-modal fundation models on instruction-tuning dataset; \citep{pahuja2025explorer} collect extensive high-quality web teajectories for tunning web agent.

% \cite{} try to intergrate whole webpage information into agent's context by efficiently parsing page HTML , \cite{} add Set-of-Mark (SoM) as light-weight visiual supplementary to the prompt
% 

\noindent\textbf{Tool-Use Web Agents and Tool Synthesis.}
Conventional browser-use agents interact with the web primarily through low-level atomic actions, which are often brittle and inefficient. In contrast, tools (typically APIs or programmatic functions) serve as higher-level actions: a single tool invocation can encapsulate and replace a long sequence of low-level atomic operations. Prior work has demonstrated the versatility of such abstractions for web-task automation~\citep{krishna2024paffa,wang2025inducing,jiang2026web,zhong2026actionengine}. 
For instance, \citet{song2025beyond} propose combining browser atomic actions with tool call actions, for which they manually collect an extensive set of human-curated APIs. \citet{zheng2025skillweaver} and \citet{prabhu2025walt} further extend this line of work by automatically synthesizing tools that encapsulate reusable agent trajectories. Unlike prior work, this work offers a more comprehensive and systematic understanding of tool use in web agents and provides a stronger empirical foundation for future research.

% to substantially improve web agent performance
% \cite{song2025beyond} proposes mix the browser atomic actions with tool-call actions, where they mannually collect extensive human-crafted APIs and significantly improves the web agent performance. \cite{zheng2025skillweaver} and \cite{prabhu2025walt} further automatically syntheziing tools by wraping up the resuable trajectories of web agent it self.
% However, due to limited, sometimes controversial experimental settings in prior work, our understanding of tools, especially synthesized ones, remains incomplete.\footnote{For example, \citet{zheng2025skillweaver} conclude that weaker web agents benefit more from tools, whereas \citet{prabhu2025walt} reach the opposite conclusion.} It is still unclear when tools help, how much they help, and what side effects they may introduce. 
% However, prior work conduct very limited exploration on the correclation between tool-synthezising model and tool-use models. some of them even have controversal conclusion, 

\section{Background and Experimental Setup}

% \subsection{Existing Frameworks for Web Tool-use Agents}
% \subsection{Web Tool-use Agent Frameworks}
\noindent\textbf{Web Tool-use Agent Frameworks.}
Table~\ref{tab:webarena_agent_compare_compact} summarizes several representative tool-use web agent frameworks. All of them aim to augment browser-use agents with tools for more effective web-task automation (see Appendix~\ref{appendix:frameowkr_details_tool_exp} for additional framework details and representative tool examples).
 
\begin{itemize}[leftmargin=1.8em,nosep]
\itemsep0em 
    \item \hybridcolor: \citet{song2025beyond} collect over 2,000 human-developed APIs, together with their documentation, for the websites in \webarena. These APIs are exposed as tools that web agents can invoke through REST requests.
    \item \skillweavercolor: Instead of manually collecting human-crafted tools, \citet{zheng2025skillweaver} propose a multi-agent framework to synthesize tools efficiently. They prompt agents to solve self-proposed tasks and then encapsulates reusable trajectories as functions, resulting in over 400 automatically generated tools for the \webarena~benchmark.
    % (\~88 tools per website). 
    \item \waltcolor: \citet{prabhu2025walt} further improve the quality of LLM-synthesized tools by using more robust action selectors and replacing tedious UI interaction sequences with direct URL manipulations. Instead of curating task-specific tools, \walt~targets identifying focused website functionalities, thereby producing a much smaller tool set.
    \item \webmcpcolor: \cite{webmcp2025} propose \webmcp, treating the loaded webpage as a client-side MCP-like~(Model Context Protocol) providers. The tools are implemented via in-page JavaScript and can be invoked by a browser agent. 
\end{itemize}

Table~\ref{tab:webarena_agent_compare_compact} compares these frameworks.
In this work, we focus on\waltcolor,\skillweavercolor, and\hybridcolor as our primary study targets, as they are representative open-source frameworks share the same downstream benchmark (i.e., \webarena), yet differ substantially in their tool design choices and practical characteristics.

\begin{table*}[t!]
\centering
\caption[Comparison of existing popular WebArena agent frameworks]%
{Comparison of existing popular tool-use agent frameworks for web automation. ``Avg. \# of tools'' means the average number of tools per website on \webarena~benchmark.}
\label{tab:webarena_agent_compare_compact}

\normalsize
\setlength{\tabcolsep}{3.2pt}
\renewcommand{\arraystretch}{1.22} %1.18

\resizebox{1.01\textwidth}{!}{
\begin{tabular}{l p{0.19\textwidth} p{0.265\textwidth} p{0.265\textwidth} p{0.21\textwidth}}
\toprule
\textbf{Aspect} 
& \multicolumn{1}{c}{\textbf{{\large \hybridcolor}}}
& \multicolumn{1}{c}{\textbf{{\large \skillweavercolor}}}
& \multicolumn{1}{c}{\textbf{{\large \waltcolor}}}
& \multicolumn{1}{c}{\textbf{{\large \webmcpcolor}}} \\
\midrule
% \cmidrule(lr){2-5}

% Action space
% & \makecell[c]{Browsing \& tool-use}
% & \makecell[c]{Browsing \& tool-use}
% & \makecell[c]{Browsing \& tool-use}
% & \\

Web observation 
& \makecell[c]{Axtree + DOM}
& \makecell[c]{Axtree + Screenshot}
& \makecell[c]{DOM + Screenshot}
& \makecell[c]{--} \\
\addlinespace[2pt]
% \cmidrule(lr){2-5}
% \midrule

Tool source
& \makecell[c]{Human}
& \makecell[c]{LLM-synthesized\\ (GPT-4o)}
& \makecell[c]{LLM-synthesized\\ (GPT-5-mini)}
& \makecell[c]{Human} \\

\addlinespace[6pt]
% \cmidrule(lr){2-5}
% \midrule

Tool format
& \makecell[c]{{REST} APIs\\ (HTTP requests)}
& \makecell[c]{{Python} Functions\\ (via Playwright)}
& \makecell[c]{{JSON} UI action flow\\ (via Playwright)}
& \makecell[c]{JS Callback APIs \\ (site built-in)} \\

\addlinespace[6pt]
% \cmidrule(lr){2-5}

Tool UI selector
& \makecell[c]{--}
& \makecell[c]{A11y role selector \\ (semantic role)}
& \makecell[c]{DOM selector + Xpath \\ + ElementHash}
& \makecell[c]{--} \\

\addlinespace[2.5pt]
% \cmidrule(lr){2-5}

Avg. \# of tools
& \makecell[c]{437}
& \makecell[c]{88}
& \makecell[c]{8}
& \makecell[c]{--} \\

\bottomrule
\end{tabular}
}
\end{table*}
% \footnotetext{test.}

% \subsection{Evaluation Benchmarks and Backbone Models}
\noindent\textbf{Evaluation Benchmarks and Backbone Models.}
%
% To draw reliable conclusions and provide a solid empirical basis for future tool-use web agent design, we conduct experiments on two widely adopted benchmarks, \webarena~\citep{zhouwebarena} and \visualwebarena~\citep{koh2024visualwebarena}, under a comprehensive and diagnostic evaluation setting.
 We conduct experiments on two widely adopted benchmarks, \webarena~\citep{zhouwebarena} and \visualwebarena~\citep{koh2024visualwebarena}, under a comprehensive and diagnostic evaluation setting.
\webarena~is a sandbox-based benchmark that simulates real-world web environments. It contains tasks across five websites: Shopping (187), CMS (182, a.k.a.\ Shopping Admin), GitLab (180), Reddit (106), and Map (109).\footnote{We exclude multi-site tasks to ensure a fair comparison, as the tools in \skillweaver~are specifically designed for single-site tasks and may not be compatible with such settings.} These tasks cover information seeking, web navigation, and web interaction, and are evaluated by human-crafted scripts using exact/fuzzy string matching or HTML parsing. \visualwebarena~is a vision-extended benchmark built on the similar websites as \webarena but with a distinct set of newly constructed tasks (e.g., visually grounded tasks). It covers three websites: Classifieds (234), Shopping (466), and Reddit (210). Both benchmarks use binary success rate as the metric.

Unlike prior work, which typically adopts either a single backbone model or homogeneous model family, we test each framework with a more diverse set of backbone models spanning different model families and scales, including \texttt{GPT-5}, \texttt{GPT-5-mini}, \texttt{GPT-5-nano}, \texttt{Grok-4.1-Fast-Reasoning}, and \texttt{Mistral-Large-3.1} (see~\ref{appendix:implementation_details} for more details).

% Both benchmarks evaluate agent's  by using human-craft scripts (answer URL)

% \section{What's the Best Practice for Tool-use Web Agents?}
\section{Experiments and Analyses}
\label{sec:experiment_and_analyses}

% ====== WALT ablation table ======
\begin{table*}[t!]
\centering
\caption{\protect\waltcolor's performances (success rate \%) with/without tools across models.}
\label{tab:walt_res}

% compact knobs
\setlength{\tabcolsep}{3.2pt}
\renewcommand{\arraystretch}{1.08}

\resizebox{1.01\textwidth}{!}{%
\begin{tabular}{
  C{1.65em}|  % (Fix A) wider tool-label column to avoid overlap
  L{2.1cm}
  R{0.95cm}R{0.85cm}R{0.85cm}R{0.85cm}R{0.85cm}>{\columncolor{AVGGray}}R{1.15cm}
  R{0.95cm}R{0.85cm}R{1.10cm}>{\columncolor{AVGGray}}R{1.15cm}
}
\toprule
& & \multicolumn{6}{c}{\GroupHdrFont \textit{WebArena}} & \multicolumn{4}{c}{\GroupHdrFont \textit{VisualWebArena}} \\
\midrule
& \ColHdrFont Models
& \multicolumn{1}{c}{\ColHdrFont Shop.}
& \multicolumn{1}{c}{\ColHdrFont CMS}
& \multicolumn{1}{c}{\ColHdrFont Gitlab}
& \multicolumn{1}{c}{\ColHdrFont Reddit}
& \multicolumn{1}{c}{\ColHdrFont Map}
& \multicolumn{1}{c}{\AVGHdrFont Avg.} 
& \multicolumn{1}{c}{\ColHdrFont Shop.}
& \multicolumn{1}{c}{\ColHdrFont Reddit}
& \multicolumn{1}{c}{\ColHdrFont Class.}
& \multicolumn{1}{c}{\AVGHdrFont Avg.} \\
\midrule

% ===================== w/o tools =====================
\multirow[c]{5}{*}{\ToolBlockLabelH{w/o tools}{7.0em}}
& GPT-5
& {43.2} & {52.2} & {59.2} & {47.4} & {62.4} & \AVGCellPlain{52.9}
& {55.8} & {42.4} & {60.3} & \AVGCellPlain{52.8} \\

& GPT-5-mini
& 39.6 & 52.7 & 43.9 & 38.6 & 56.0 & \AVGCellPlain{46.2}
& 43.6 & 38.1 & 56.4 & \AVGCellPlain{46.0} \\

& GPT-5-nano
& 21.9 & 27.5 & 23.0 & 17.5 & 32.1 & \AVGCellPlain{24.4}
& 24.9 & 14.3 & 32.5 & \AVGCellPlain{23.9} \\

& Grok-4.1-R
& 37.0 & 48.9 & 46.4 & 35.1 & 55.0 & \AVGCellPlain{44.5}
& 45.1 & 36.7 & 50.0 & \AVGCellPlain{43.9} \\

& Mistral-L-3.1
& 25.0 & 43.4 & 39.8 & 28.1 & 40.4 & \AVGCellPlain{35.3}
& 39.9 & 33.8 & 41.9 & \AVGCellPlain{38.5} \\

\midrule

% ===================== w/ tools (Fix B: rowcolors block shading, more stable) =====================
\rowcolors{1}{ToolBlockGray}{ToolBlockGray} % same color every row in this block
\multirow[c]{5}{*}{\ToolBlockLabelH{w/ tools}{7.0em}}
& GPT-5
& {42.7} & {57.1} & {54.6} & {39.5} & {60.6} & \AVGCellDelta{50.9}{\MydownArrow{2.0}}
& {53.4} & {38.6} & {63.7} & \AVGCellDelta{51.9}{\MydownArrow{0.9}} \\

& GPT-5-mini
& 40.6 & 55.5 & 42.9 & 33.3 & 57.8 & \AVGCellDelta{46.0}{\MydownArrow{0.2}}
& 49.4 & 34.3 & 59.0 & \AVGCellDelta{47.6}{\MyUpArrow{1.6}} \\

& GPT-5-nano
& 23.4 & 34.1 & 22.4 & 27.2 & 35.8 & \AVGCellDelta{28.6}{\MyUpArrow{4.2}}
& 32.2 & 21.4 & 39.3 & \AVGCellDelta{31.0}{\MyUpArrow{7.1}} \\

& Grok-4.1-R
& 34.9 & 49.5 & 48.0 & 43.9 & 56.9 & \AVGCellDelta{46.6}{\MyUpArrow{2.1}}
& 42.5 & 40.0 & 51.7 & \AVGCellDelta{44.7}{\MyUpArrow{0.8}} \\

& Mistral-L-3.1
& 29.7 & 48.4 & 40.3 & 36.8 & 45.9 & \AVGCellDelta{40.2}{\MyUpArrow{4.9}}
& 43.1 & 33.3 & 46.6 & \AVGCellDelta{41.0}{\MyUpArrow{2.5}} \\

% \rowcolors{1}{}{} % reset
\bottomrule
\end{tabular}%
} % end resizebox

\end{table*}

% ====== Skillweaver table ======
\begin{table*}[t!]
\centering
\caption{\protect\skillweavercolor's performances (success rate \%) with/without tools across models.}
\label{tab:skillweaver_res}

% compact knobs
\setlength{\tabcolsep}{3.2pt}
\renewcommand{\arraystretch}{1.08}

\resizebox{1.01\textwidth}{!}{%
\begin{tabular}{
  C{1.65em}|  % (Fix A) wider tool-label column to avoid overlap
  L{2.1cm}
  R{0.95cm}R{0.85cm}R{0.85cm}R{0.85cm}R{0.85cm}>{\columncolor{AVGGray}}R{1.15cm}
  R{0.95cm}R{0.85cm}R{1.10cm}>{\columncolor{AVGGray}}R{1.15cm}
}
\toprule
& & \multicolumn{6}{c}{\GroupHdrFont \textit{WebArena}} & \multicolumn{4}{c}{\GroupHdrFont \textit{VisualWebArena}} \\
\midrule
& \ColHdrFont Models
& \multicolumn{1}{c}{\ColHdrFont Shop.}
& \multicolumn{1}{c}{\ColHdrFont CMS}
& \multicolumn{1}{c}{\ColHdrFont Gitlab}
& \multicolumn{1}{c}{\ColHdrFont Reddit}
& \multicolumn{1}{c}{\ColHdrFont Map}
& \multicolumn{1}{c}{\AVGHdrFont Avg.} 
& \multicolumn{1}{c}{\ColHdrFont Shop.}
& \multicolumn{1}{c}{\ColHdrFont Reddit}
& \multicolumn{1}{c}{\ColHdrFont Class.}
& \multicolumn{1}{c}{\AVGHdrFont Avg.} \\
\midrule

% ===================== w/o tools =====================
\multirow[c]{5}{*}{\ToolBlockLabelH{w/o tools}{7.0em}}
& GPT-5
& {30.5} & {43.4} & {42.2} & {31.1} & {48.6} & \AVGCellPlain{39.2}
& 41.2 & 31.0 & 41.4 & \AVGCellPlain{37.9} \\

& GPT-5-mini
& 22.5  & 25.3 & 23.9 & 21.7 & 36.7 & \AVGCellPlain{26.0}
& 29.6 & 21.4 & 26.7 & \AVGCellPlain{25.9} \\

& GPT-5-nano
& 15.5 & 9.9 & 11.1 & 7.5 & 11.0 & \AVGCellPlain{11.0}
& 13.3 & 10.5 & 9.91 & \AVGCellPlain{11.2} \\

& Grok-4.1-R
& 32.1 & 30.8 & 26.1 & 31.1 & 44.0 & \AVGCellPlain{32.8}
& 38.4 & 28.1 & 33.2 & \AVGCellPlain{33.2} \\

& Mistral-L-3.1
& 21.9 & 25.8 &  25.0 & 16.0 & 29.4 & \AVGCellPlain{23.6}
& 25.3 & 14.3 & 27.2 & \AVGCellPlain{22.3} \\

\midrule

% ===================== w/ tools (Fix B: rowcolors block shading, more stable) =====================
\rowcolors{1}{ToolBlockGray}{ToolBlockGray} % same color every row in this block
\multirow[c]{5}{*}{\ToolBlockLabelH{w/ tools}{7.0em}}
& GPT-5
& {28.3} & {42.9} & {44.4} & {30.2} & {41.3} & \AVGCellDelta{37.4}{\MydownArrow{1.8}}
& 40.3 & 28.1 & 34.9  & \AVGCellDelta{34.4}{\MydownArrow{3.5}} \\

& GPT-5-mini
&23.5 & 28.6 & 29.4 & 25.5 & 29.4 & \AVGCellDelta{27.3}{\MyUpArrow{1.3}}
& 27.7 & 22.9 & 22.4 & \AVGCellDelta{24.3}{\MydownArrow{1.6}} \\

& GPT-5-nano
& 10.7 & 12.1 & 13.3 & 11.3 & 18.3 & \AVGCellDelta{13.1}{\MyUpArrow{2.1}}
& 15.2 & 13.3 & 11.2 & \AVGCellDelta{13.2}{\MyUpArrow{2.0}} \\

& Grok-4.1-R
& 27.3 & 29.1 & 32.2 & 27.4 & 35.8 & \AVGCellDelta{30.4}{\MydownArrow{2.4}}
& 35.4 & 23.8 & 29.7 & \AVGCellDelta{29.6}{\MydownArrow{3.6}} \\

& Mistral-L-3.1
& 20.9 & 26.4 & 21.7 & 17.9 & 27.5 & \AVGCellDelta{22.9}{\MydownArrow{0.7}}
& 23.0 & 17.1 & 21.6 & \AVGCellDelta{20.6}{\MydownArrow{1.7}} \\

% \rowcolors{1}{}{} % reset
\bottomrule
\end{tabular}%
} % end resizebox

\end{table*}

% ====== Hybrid-Agent table ======
\begin{table*}[t!]
\centering
\caption{\protect\hybridcolor's performances (success rate \%) with/without tools across models.}
\label{tab:hybrid_res}

% compact knobs
\setlength{\tabcolsep}{3.2pt}
\renewcommand{\arraystretch}{1.08}

\resizebox{1.01\textwidth}{!}{%
\begin{tabular}{
  C{1.65em}|  % (Fix A) wider tool-label column to avoid overlap
  L{2.1cm}
  R{0.95cm}R{0.85cm}R{0.85cm}R{0.85cm}R{0.85cm}>{\columncolor{AVGGray}}R{1.15cm}
  R{0.95cm}R{0.85cm}R{1.10cm}>{\columncolor{AVGGray}}R{1.15cm}
}
\toprule
& & \multicolumn{6}{c}{\GroupHdrFont \textit{WebArena}} & \multicolumn{4}{c}{\GroupHdrFont \textit{VisualWebArena}} \\
\midrule
& \ColHdrFont Models
& \multicolumn{1}{c}{\ColHdrFont Shop.}
& \multicolumn{1}{c}{\ColHdrFont CMS}
& \multicolumn{1}{c}{\ColHdrFont Gitlab}
& \multicolumn{1}{c}{\ColHdrFont Reddit}
& \multicolumn{1}{c}{\ColHdrFont Map}
& \multicolumn{1}{c}{\AVGHdrFont Avg.} 
& \multicolumn{1}{c}{\ColHdrFont Shop.}
& \multicolumn{1}{c}{\ColHdrFont Reddit}
& \multicolumn{1}{c}{\ColHdrFont Class.}
& \multicolumn{1}{c}{\AVGHdrFont Avg.} \\
\midrule

% ===================== w/o tools =====================
\multirow[c]{5}{*}{\ToolBlockLabelH{w/o tools}{7.0em}}
& GPT-5
& {25.1} & {39.6} & {41.7} & {38.7} & {37.6} & \AVGCellPlain{36.5}
& {33.9} & {30.1} & {--} & \AVGCellPlain{32.0} \\

& GPT-5-mini
& 15.0 & 24.2 & 19.4 & 15.1 & 22.9 & \AVGCellPlain{19.3}
& 18.7 & 13.3 & -- & \AVGCellPlain{16.0} \\

& GPT-5-nano
& 8.6 & 12.1 & 9.4 & 9.4 & 11.0 & \AVGCellPlain{10.1}
& 10.4 & 7.3 & -- & \AVGCellPlain{8.9} \\

& Grok-4.1-R
& 19.3 & 27.5 & 23.9 & 29.2 & 31.2 & \AVGCellPlain{26.2}
& 25.7 & 25.3 & -- & \AVGCellPlain{25.5} \\

& Mistral-L-3.1
& 12.8 & 19.8 & 18.9 & 11.3 & 20.2 & \AVGCellPlain{16.6}
& 13.3 & 10.1 & -- & \AVGCellPlain{11.7} \\

\midrule

% ===================== w/ tools (Fix B: rowcolors block shading, more stable) =====================
\rowcolors{1}{ToolBlockGray}{ToolBlockGray} % same color every row in this block
\multirow[c]{5}{*}{\ToolBlockLabelH{w/ tools}{7.0em}}
& GPT-5
& {33.2} & {50.5} & {55.0} & {57.5} & {53.2} & \AVGCellDelta{49.9}{\MyUpArrow{13.4}}
& {37.6} & {33.2} & {--} & \AVGCellDelta{35.4}{\MyUpArrow{3.4}} \\

& GPT-5-mini
& 25.1 & 37.9 & 46.1 & 42.5 & 43.1 & \AVGCellDelta{38.9}{\MyUpArrow{19.6}}
& 26.5 & 21.9 & -- & \AVGCellDelta{24.2}{\MyUpArrow{8.2}} \\

& GPT-5-nano
& 13.4 & 23.1 & 25.6 & 30.2 & 20.2 & \AVGCellDelta{22.5}{\MyUpArrow{12.4}}
& 15.8 & 12.9 & -- & \AVGCellDelta{14.4}{\MyUpArrow{5.5}} \\

& Grok-4.1-R
& 27.3 & 36.3 & 49.4 & 54.7 & 45.9 & \AVGCellDelta{42.7}{\MyUpArrow{16.5}}
& 30.2 & 28.8 & -- & \AVGCellDelta{29.5}{\MyUpArrow{4.0}} \\

& Mistral-L-3.1
& 21.4 & 28.0 & 33.3 & 37.7 & 35.8 & \AVGCellDelta{31.2}{\MyUpArrow{14.6}}
& 19.3 & 15.9 & -- & \AVGCellDelta{17.6}{\MyUpArrow{5.9}} \\

% \rowcolors{1}{}{} % reset
\bottomrule
\end{tabular}%
} % end resizebox

\end{table*}

Prior work has shown that tool use is highly versatile in web domain, but existing evidence is often limited to individual frameworks or incomplete evaluation settings. As a result, it remains unclear how effective tools truly are, when they help versus fail, what costs they introduce, and what design principles should guide future tool-use web agents. In this section, we conduct comprehensive empirical studies across aforementioned frameworks, backbone models, and benchmarks, where we try to answer the following questions:
% with the goal of developing a more systematic understanding of tool use in web agents
\begin{itemize}[leftmargin=1.8em,nosep]
\itemsep0em 
    \item $\mathcal{Q}_1$: Do tools always yield consistent performance gains? How effective are they, and when might they fail?~(\cref{subsec:tool_cannot_always_gains} \& \cref{subsec:tool_as_capacity_distil})
    
    \item $\mathcal{Q}_2$: Are more comprehensive and complex tools necessarily better? (\cref{subsec:tool_cannot_over_complex})
    
    \item $\mathcal{Q}_3$: What is the effective tool design principle? (\cref{subsec:tool_coverage})
    
    \item $\mathcal{Q}_4$: Does tool use truly improve the web-task execution efficiency of agents? (\cref{subsec:tax_of_tool_efficiency})
    
    \item $\mathcal{Q}_5$: How do semantic skills compare with programmatic tools? (\cref{subsec:skill_vs_tool})
    
    \item $\mathcal{Q}_6$: Does webpage vision remain useful in the tool-use setting? (\cref{subsec:vision_benefits})
\end{itemize}

\subsection{Synthetic Tools Do Not Always Yield Consistent Gains}
\label{subsec:tool_cannot_always_gains}
% Previous works conlucded that, tools can be effective for enhancing web task aacuracy compared with pure browser agent, including those tools synthezed by agent itself. However, there is limited tool quality quantitive study,
% will llm-synthetic tools consistently bring benefis when the model scaling? Will the quality becomes issues for synthetic tools, especially when the browser-use agent is really strong enough, do we really still need tools?
% it's still unclear that how much benefits can the tool brings to different models, and how the tool quality will impact the performance, 
Prior work has generally portrayed tools as uniformly beneficial for web browser agents, including both human-designed tools and agent-synthesized ones~\citep{krishna2024paffa,song2025beyond,zheng2025skillweaver,prabhu2025walt}. Yet these claims are typically based on relatively limited evaluations and lack sufficiently comprehensive quantitative support. 
% Consequently, it remains unclear how much benefit tools (especially synthesized tools) actually bring to backbone models of different scales, whether such gains remain consistent as browser agents themselves become stronger, and how sensitive performance is to tool quality. 
% In particular, once the base browser agent is already highly capable, it is no longer obvious that tools are always necessary; poor-quality synthesized tools may instead introduce additional overhead, instability, or failure modes.

% To quantify the gains brought by tools, we evaluate backbone models spanning multiple scales and families, across three aforementioned frameworks. 
Tables~\ref{tab:walt_res}, \ref{tab:skillweaver_res}, and \ref{tab:hybrid_res} quantify the gains brought by tools in a comprehensive setting spanning three frameworks and five backbone models.
First,\hybridcolor consistently improves after tool use is enabled on both benchmarks. We attribute this to the high reliability of its website-native REST APIs, which cover a broad range of website functionalities, whereas the LLM-synthesized tools in \walt~and~\skillweaver~lead to more mixed outcomes.

% Specifcially, \walt's tools is constructed by GPT-5-mini, and GPT-5-mini occupies the second base browser agent performance (See the upper part of Table~\ref{tab:walt_res}), as a results, all the backbone models whose baseline performance lower than GPT-5-mini, get performance gains after using GPT-5-mini synthetic tools. While GPT-5-mini itself gets comparable performance after usiing it's self-crafted tools, and GPT-5's performance has generalyy drop.
% Similar observations can be found in \skillweaver's results. Since the tools of \skillweaver~is constructed by GPT-4o (as reported by \citep{zheng2025skillweaver}, GPT-4o w/o tools has 22.6 average performance on \webarena), in Table~\ref{tab:skillweaver_res}, only GPT-5-nano whose baseline performance is lower than GPT-4o get consistent performance gains, while all other models w/tools performance drops / comparable as w/o tools baseline.
Specifically, the tools used by\waltcolor are synthesized by GPT-5-mini. In the \textit{w/o tools} block of Table~\ref{tab:walt_res}, {GPT-5-mini} is the second-strongest backbone model, behind GPT-5 but stronger than all others. Correspondingly, in the \textit{w/ tools} block, all models whose baseline performance is lower than GPT-5-mini consistently benefit from these GPT-5-mini-synthesized tools.
% including GPT-5-nano, Grok-4.1-R, and Mistral-L-3.1 on both \webarena~and \visualwebarena.
By contrast, GPT-5-mini itself shows mixed or marginal gains, while GPT-5 generally degrades after tool use. 
% This suggests that the gains from synthesized tools are not uniform, but instead depend strongly on the relative capability gap between the tool-generating model and the downstream backbone model.
A similar pattern appears in Table~\ref{tab:skillweaver_res}. The tools in\skillweavercolor are synthesized by GPT-4o, whose reported \webarena~performance in \citet{zheng2025skillweaver} is 22.6\% (w/o tools). Comparing this reference point with the \textit{w/o tools} block of Table~\ref{tab:skillweaver_res}, only GPT-5-nano falls clearly below the GPT-4o baseline. Consistently, GPT-5-nano is also the only model that shows stable gains from tool use across both benchmarks. All other backbones either become worse or exhibit only mixed changes after adopting tools. 

These observations suggest that LLM-synthesized tools yield consistent gains only when the ``tool user'' is clearly weaker than the ``tool developer''.\footnote{Here, ``weaker'' and ``stronger'' are framework-dependent rather than absolute notions. For example, Grok-4.1-R is stronger than GPT-5-mini in Table~\ref{tab:skillweaver_res}, but the ordering is reversed in Table~\ref{tab:walt_res}.}

\subsection{Tool Synthesis as Capability Distillation: Strong-to-Weak Transfer}
\label{subsec:tool_as_capacity_distil}

\begin{table}[t!]
% \begin{wraptable}[10]{R}{0.48\textwidth}
\centering
% \scriptsize
\normalsize
\caption{Results of varying the tool constructor.}
\label{tab:tool_created_by_stronger}

\begin{tabular}{lrrr}
\toprule
                     & \multicolumn{1}{c}{GPT-5} & \multicolumn{1}{c}{GPT-5-mini} & \multicolumn{1}{c}{GPT-5-nano} \\ \midrule

\rowcolor{LightGrayRow}
\textbf{\skillweaver}            & 43.4                      & 25.3                           & 9.9                            \\ 
% \midrule
\hspace{0.8em}w/ tools (by \texttt{GPT-4o}) & {\textcolor{OrangeRed}{$\bm\downarrow$}}~42.9                  & {\textcolor{ForestGreen}{$\bm{\uparrow}$}}~28.6                       &{\textcolor{ForestGreen}{$\bm{\uparrow}$}}~12.1                       \\

\hspace{0.8em}w/ tools (by \texttt{GPT-5-mini}) & {\textcolor{OrangeRed}{$\bm\downarrow$}}~41.8                  & {\textcolor{ForestGreen}{$\bm{\uparrow}$}}~29.7                       &{\textcolor{ForestGreen}{$\bm{\uparrow}$}}~16.5                       \\

\hspace{0.8em}w/ tools (by \texttt{GPT-5})  & {\textcolor{ForestGreen}{$\bm{\uparrow}$}}~44.5                  & {\textcolor{ForestGreen}{$\bm{\uparrow}$}}~31.3                       & {\textcolor{ForestGreen}{$\bm{\uparrow}$}}~17.6                       \\ 

\midrule

\rowcolor{LightGrayRow}
\textbf{\walt}             & 47.4                      & 38.6                           & 17.5                            \\ 
% \midrule
\hspace{0.8em}w/ tools (by \texttt{GPT-4o}) & {\textcolor{OrangeRed}{$\bm\downarrow$}}~37.7                 & {\textcolor{OrangeRed}{$\bm\downarrow$}}~31.6                       &{\textcolor{ForestGreen}{$\bm{\uparrow}$}}~24.6                       \\

\hspace{0.8em}w/ tools (by \texttt{GPT-5-mini}) & {\textcolor{OrangeRed}{$\bm\downarrow$}}~39.5                  & {\textcolor{OrangeRed}{$\bm{\downarrow}$}}~33.3                       & {\textcolor{ForestGreen}{$\bm{\uparrow}$}}~27.2                       \\
\hspace{0.8em}w/ tools (by \texttt{GPT-5})  & {\textcolor{ForestGreen}{$\bm{\uparrow}$}}~48.2                  & {\textcolor{ForestGreen}{$\bm{\uparrow}$}}~43.0                       & {\textcolor{ForestGreen}{$\bm{\uparrow}$}}~29.8
\\
\bottomrule
\end{tabular}
\end{table}
% \end{wraptable}

% Table~\ref{tab:walt_res},~\ref{tab:skillweaver_res}, and~\ref{tab:hybrid_res} fix the tool developer while vary the tool user, to further enhance our previous section's conclusion, we do reverse thing: fixing tool user, while switching tool developers. 
% To this end, we use different models as tool constructors to synthezing the tools for \walt~and~\skillweaver (by following their tool construction pipline). Since tool syntheziing is expensive and time comsuming, we conduct experiments on selective websites (i.e., \skillweaver~on CMS, \walt~on Reddit, as they represent distinct performance distribution before/after enabling tools).
% Table~\ref{tab:tool_created_by_stronger} shows the results. Accordingly, only GPT-5-nano, which is significantly weaker than all the tool constructors, get consistent improvement; while  GPT-5-mini obtains consistent gains across two frameworks only when the tool constructor is GPT-5.
Tables~\ref{tab:walt_res}, \ref{tab:skillweaver_res}, and \ref{tab:hybrid_res} primarily study scaling the backbone models that use tools, while keeping the tool developer fixed. To further strengthen our conclusion, we additionally study the complementary scaling direction: varying the models used to construct tools while fixing the tool user. Specifically, we synthesize tools for \walt~and \skillweaver~(via their tool‑construction pipeline) using models of different sizes. Because tool synthesis is expensive, we restrict this analysis to two representative websites, namely CMS for \skillweaver~and Reddit for \walt, where the before/after-tool score trends are most distinct.

Table~\ref{tab:tool_created_by_stronger} shows a consistent pattern: GPT-5-nano, which is clearly weaker than all tool constructors, is the only model that benefits consistently across all settings, whereas GPT-5-mini shows reliable gains only when the tool constructor is GPT-5.
% These results reinforce our earlier conclusion that LLM-synthesized tools primarily help clearly weaker models, while their benefits for stronger or comparable models are limited and inconsistent.
{These observations further supports the view that model-synthesized tools behave like a form of capability distillation from stronger models to weaker ones, but the reverse does not necessarily hold.
}

% \textcolor{red}{We therefore interpret LLM-based tool synthesis as a form of capability distillation. Rather than transferring domain knowledge alone, it externalizes the teacher model’s general web browsing competence into callable programs, which can then function as ``programmatic memory'' for the target website and be reused by weaker student models.
% }
% memory, knowledge distillation, 
% tool is more like procedural, programatic 

\subsection{Tools Should Not Be Overly Complex: Decoupling Agentic Reasoning from Tools}
\label{subsec:tool_cannot_over_complex}

% Since tools can act as external programmatic memory, 
Intuitively, a sufficiently comprehensive tool could solve a complex web task end-to-end with little or no additional browser interaction. This raises a natural question: are more complex tools always better?
To study this, we categorize tool complexity into three levels using LLM prompting and heuristic rules (more details in Appendix~\ref{appendix:tool_complexity_level}): \textbf{high}, \textbf{medium}, and \textbf{low}. High-level tools encapsulate long and task-specific interaction sequences~(e.g., \texttt{search\_and\_mark\_onsale}); medium‑level tools correspond to more focused and generalizable intentions~(e.g., \texttt{search}); low-level tools are short, atomic operations~(e.g., \texttt{navigate}).
%  with strong UI-state dependence

\begin{figure}
% \vspace{-2.3em}
\setlength{\abovecaptionskip}{-10pt}
\begin{center}
\centering
\includegraphics[width=0.95\textwidth]{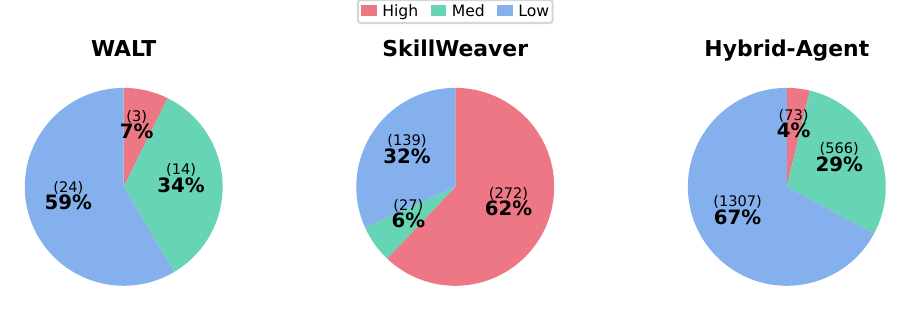}
\end{center}

\caption{Distribution of tool complexity levels across the three frameworks.
}
\label{fig:tool_complexity_pie}
\end{figure}

\begin{figure}
% \vspace{-2.3em}
\setlength{\abovecaptionskip}{-10pt}
\begin{center}
\centering
\includegraphics[width=0.95\textwidth]{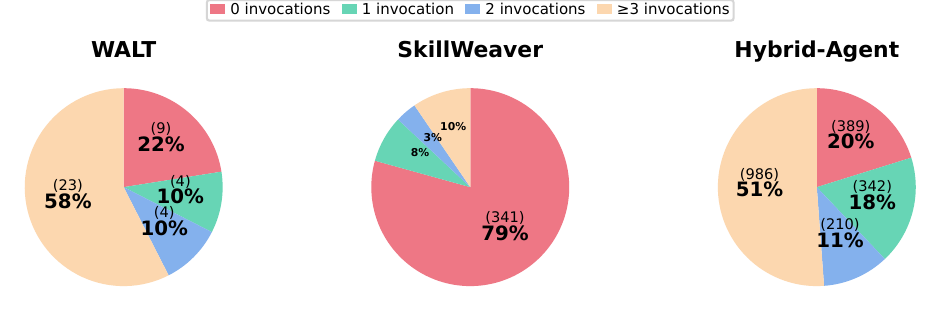}
\end{center}

\caption{Distribution of tool invocations. ``X invocations'' indicates that the tool is invoked in X tasks of \webarena, e.g., ``0 invocations'' means the tools are \textbf{unused} across all tasks.
}
\label{fig:tool_invocation}
\end{figure}

% Figure~\ref{fig:tool_complexity_pie} shows that most human-developed tools in \hybrid~are of low or medium complexity, with only a very small fraction being high-level. \walt~exhibits a similar distribution. By contrast, \skillweaver~contains a substantial number of high-complexity tools. 
% % Furthermore, 
% Due to agreesively atempting for wrapping overly deep trajecotry, we find that most tools of \skillweaver~are overly task-specific and involve complex control logic (e.g., loops/conditions), which makes them strongly dependent on particular UI states and reduces their generalizability and utility. for example, Figure~\ref{fig:tool_invocation} exhibts te tool invocation distribution, where the extnsive tools in \skillweaver~are nevered invoked during evluaating on \webarena.
Figure~\ref{fig:tool_complexity_pie} shows that most human-developed tools in \hybridcolor are of low or medium complexity, with only a small fraction being high-level. \waltcolor exhibits a similar distribution. By contrast, \skillweavercolor contains a substantially larger proportion of high-complexity tools. We find that, by aggressively attempting to encapsulate deep trajectories, many \skillweaver's~tools involve complex control logic (e.g., loops/conditionals) and lead to strong UI-state dependence. This makes them overly task-specific, reducing both their generalizability and their practical utility. This consequence is further reflected in Figure~\ref{fig:tool_invocation}, which shows that a large portion of \skillweaver‘s tools are never invoked during evaluation on \webarena. This further suggests that scaling the tool set with overly complex tools does not translate into proportional practical utility~(i.e., ineffective scaling).

We therefore conclude that more complex tools are not necessarily better, and may instead reduce generalizability and practical utility. Moreover, complex tools are often overly aggressive in attempting to encapsulate entire tasks, thereby crowding out reasoning and planning that should be handled more flexibly by the browser agent. A better design principle is to \textbf{decouple agentic reasoning from tools}: tools should encapsulate only \textbf{deterministic} and reusable UI operations to ensure efficiency and reliable replay, while the agent should retain \textbf{context-sensitive} reasoning and planning based on real-time grounding from the current web state. In this sense, effective web agents should not rely on tools exclusively; instead, a better design principle is to maintain a balanced mixture of tool use and direct browser interaction~\citep{song2025beyond}.

% \vspace{-3.3em}

\subsection{The Essence of Tool Design: Functional Coverage and Composition}
\label{subsec:tool_coverage}

\begin{table}[t!]
\centering
\setlength{\abovecaptionskip}{-0.5pt}
\setlength{\belowcaptionskip}{1.8pt}
\caption{Distribution of \webarena~tasks by the number of tools involved, together with their success rates. ``\(\geq\)2 tools'' indicates that agent combines multiple tools to solve a task.}
\label{tab:tool_combo_usage}

\normalsize
\setlength{\tabcolsep}{7.5pt}
\renewcommand{\arraystretch}{1.08}

\begin{tabular}{l c >{\columncolor{LightGrayRow}}c c >{\columncolor{LightGrayRow}}c c >{\columncolor{LightGrayRow}}c}
\toprule
& \multicolumn{2}{c}{\textbf{\textit{\large \walt}}}
& \multicolumn{2}{c}{\textbf{\textit{\large \skillweaver}}}
& \multicolumn{2}{c}{\textbf{\textit{\large \hybrid}}} \\
\cmidrule(lr){2-3} \cmidrule(lr){4-5} \cmidrule(lr){6-7}
& \multicolumn{1}{c}{\textbf{Task Num.}}
& \multicolumn{1}{c}{\textbf{Acc.}}
& \multicolumn{1}{c}{\textbf{Task Num.}}
& \multicolumn{1}{c}{\textbf{Acc.}}
& \multicolumn{1}{c}{\textbf{Task Num.}}
& \multicolumn{1}{c}{\textbf{Acc.}} \\
\midrule

% \rowcolor{LightGrayRow}
\textbf{0 tools} & 357~{\tiny (45\%)} & 56.3\% & 159~{\tiny (21\%)} & 42.1\% & 82~{\tiny (11\%)} & 39.0\% \\

% \rowcolor{LightGrayRow}
\textbf{1 tool} & 291~{\tiny (37\%)} & 43.6\% & 408~{\tiny (54\%)} & 39.2\% & 286~{\tiny (37\%)} & 50.3\% \\

% \rowcolor{LightGrayRow}
\textbf{\(\geq\)2 tools} & 140~{\tiny (18\%)} & 32.1\% & 194~{\tiny (25\%)} & 27.5\% & 396~{\tiny (52\%)} & 41.2\% \\

\bottomrule
\end{tabular}
\end{table}

In \cref{subsec:tool_cannot_over_complex}, we conclude that the tools should not be overly complex. Ideally, even when a single tool cannot solve a complex web task end-to-end, an agent may still achieve the task by combining multiple medium- or low-level tools. We therefore further analyze tool compositionality and functional coverage.

Table~\ref{tab:tool_combo_usage} shows the distribution of \webarena~tasks by the number of unique tools involved, grouped into three categories: 0 tools, 1 tool, and 
\(\geq\)2 tools. Notably, the human-developed tools in \hybrid~exhibit much stronger compositionality: a larger proportion of tasks are solved through multi-tool composition, and these cases also achieve higher success rates than in the other two frameworks. In contrast, \walt~and \skillweaver, both based on LLM-synthesized tools, show weaker compositionality, with both lower rates of multi-tool usage and lower success rates for multi-tool tasks.
% Table~\ref{tab:tool_combo_usage} shows the distribution of \webarena~tasks by the (unique) number of involved tools, grouping tasks into three categories: 0 tools, 1 tool, and \(\geq\)2 tools. Notably, the human-developed tools in \hybrid~exhibit substantially stronger compositionality: a much larger proportion of tasks are solved through multi-tool composition, and these multi-tool trajectories also achieve higher success rates than those of the other two frameworks. In contrast, \walt~and \skillweaver, both relying on LLM-synthesized tools, show weaker tool compositionality. 
% reflected in both a lower proportion of multi-tool usage and lower success rates when multiple tools are involved.

We further observe that only a small fraction of tasks in \hybrid~are solved with pure browser actions, suggesting that its tools provide broad functional coverage over \webarena~task intentions. In other words, the agent can find suitable tools for most tasks. By contrast, \walt~shows the opposite trend, with a much larger portion of tasks still relying on pure browser interaction, indicating more limited coverage of its tool set.

We therefore draw the following high-level design principle: neither tool complexity nor tool-set size alone leads to proportional improvements in web-task performance. The more fundamental factor is coverage of website functionality through composable tools. Tools do not always need to solve tasks end-to-end in one shot; lower-level tools that remain close to atomic browser actions are acceptable, provided that they compose reliably and can jointly serve higher-level user intentions. From this perspective, effective tool design is less about maximizing per-tool comprehensiveness and more about enabling broad coverage of frequent real-world task intentions through composition.

% Ideally, a single tool might not be able to accomplish a ocmplex web task end to end, but the agent can combine multiple medium- or low-level tools to achive higher-level task intentions. Therefore, we further prob the tool combinationlity and web site funcitonnality coverage.

% Table~\ref{tab:tool_combo_usage} presents the task distirbution where How many tools involved in resolving the tasks in \webarena.  concretely, we categorize \webarena's tasks into 0 tools, 1 tools, and \(\geq\)2 tools involved in resolving the tasks. Notably, the human-developed tools in \hybrid~exhibit high combinationnality, where the most tasks are accomplished by \hybrid~via multiple tool combination and also reaches relatively higher success rate comes from tool combination compared with the other two frameworks, while the other two frameworks with llm-synthetic tools presents a relatively low combinationality (lower proprtion of tool combination and lower tool combination accuracy). Furthermore, we also observe only a small portion of tasks in \hybrid solved by soley borwsering actions, which also implies the high coverage of those tools in \hybrid~that uniformly covers the task intentions from \webarena. so agent can find approporate tools for most tasks.  While \walt~shows reversed trend, where agent still really on pure bowsering action for large part of tasks.

% as we conclude that the tool cannot be overlly complex, To furtuher charactrize the effectives tool, we investigate the combinationilty of tools in exsting frameworks 

\subsection{The Tax of Tools: Token Cost and Action Overhead}
\label{subsec:tax_of_tool_efficiency}

\begin{figure}[!t]
% \vspace{-2.3em}
\setlength{\abovecaptionskip}{-5pt}
\begin{center}
\centering
\includegraphics[width=1.0\textwidth]{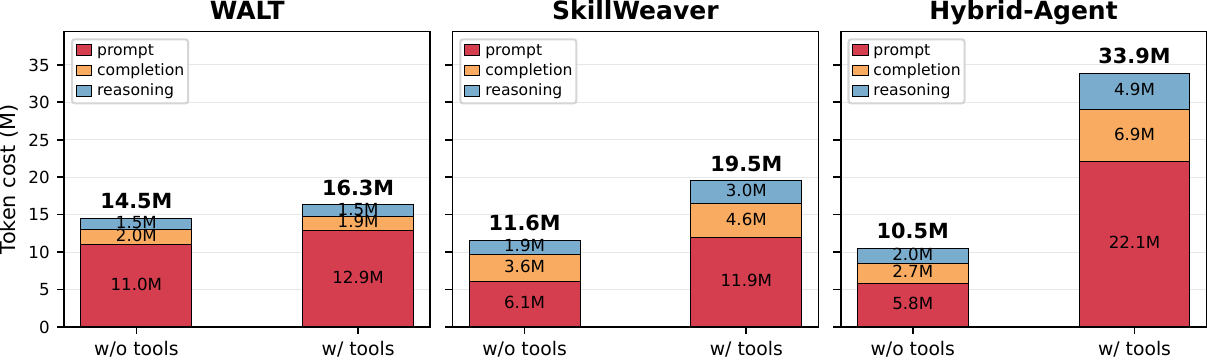}
\end{center}

\caption{Average token cost per website on \webarena.
}
\label{fig:token_cost}
\end{figure}

\begin{figure}[!t]
% \vspace{-2.3em}
\setlength{\abovecaptionskip}{-5pt}
\begin{center}
\centering
\includegraphics[width=0.5\textwidth]{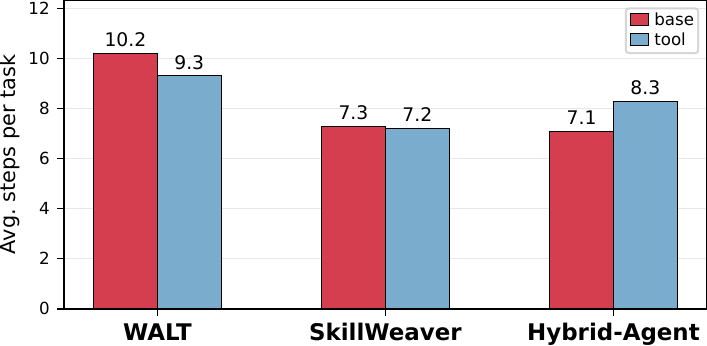}
\end{center}

\caption{Average number of agent steps required to complete a task on \webarena.
}
\label{fig:avg_agent_step}
\end{figure}

% Tools can be viewed as wrappers of reusable web-use experience distilled from teacher models, and thus may ideally provide shortcuts for browser agents. 
The availability of tools allows web agents to take ``shortcuts'' instead of relying solely on tedious browser interactions.
% , which often incur long latency
To examine whether this advantage translates into actual efficiency gains, we further analyze tool-use agents from two perspectives: token cost and action overhead.

Figure~\ref{fig:token_cost} shows the token cost of the three frameworks on \webarena, decomposed into prompt, completion, and reasoning tokens. Across all frameworks, enabling tools generally increases the average token cost per website, with the increase being especially pronounced for \hybrid, largely due to its extremely large API tool library.

Figure~\ref{fig:avg_agent_step} examines action efficiency in terms of average agent steps per task. Among the three frameworks, only \walt~becomes more step-efficient after enabling tools; \skillweaver~remains roughly unchanged, while \hybrid~becomes less efficient. We attribute these differences primarily to tool-set size and tool utility. \hybrid~contains a very large tool library, forcing the agent to spend additional steps on tool retrieval, selection, and inspection. \walt, by contrast, has only a few tools per website (Table~\ref{tab:webarena_agent_compare_compact}), and these tools are mostly of low or medium complexity, making them easier for the agent to balance against direct browser actions. \skillweaver~shows mixed outcomes: while tools indeed reduce the number of steps for some tasks, their excessive complexity and task-specificity limit their practical utility and make them harder for the agent to use reliably in other cases.\footnote{In some \skillweaver~trajectories, overly task-specific tools lead to incorrect calls or incompatible usages, requiring the agent to recover from tool errors or repeatedly adjust tool-call parameters.}

These results reveal a hidden tax of tool use: tools do not necessarily make web task execution more efficient. In practice, searching through large tool libraries and handling low-utility LLM-synthesized tools can even increase both action overhead and token cost, while tool construction/synthesis itself also incurs non-trivial expense.

\subsection{Skills as a Flexible Alternative to Tools}
\label{subsec:skill_vs_tool}

% \begin{wrapfigure}{r}{0.48\columnwidth}
% \centering
% \includegraphics[width=\linewidth]{figures/paper_barplot_cms_reddit.pdf}
% \caption{CMS and Reddit results.}
% \end{wrapfigure}

% \begin{wrapfigure}[16]{R}{0.48\textwidth}
\begin{figure}
% \vspace{-2.3em}
\setlength{\abovecaptionskip}{-6pt}
\begin{center}
\centering
\includegraphics[width=0.99\textwidth]{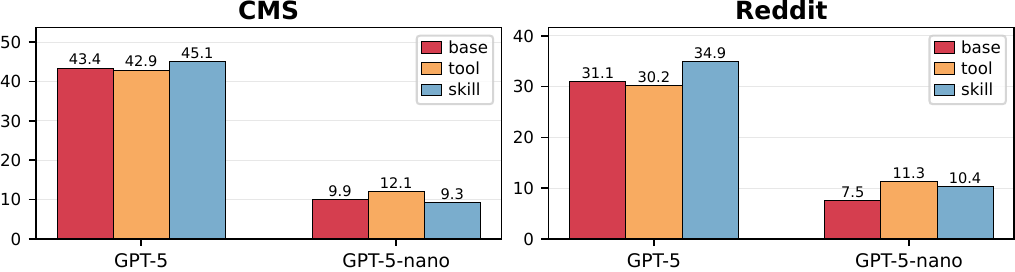}
\end{center}

\caption{Comparison of tool and skill in \skillweaver.}
\label{fig:tool_vs_skill}
\end{figure}
Tools expose executable procedures but largely remain black boxes to the agent: the model can invoke them, yet has no direct access to the internal action sequence once execution begins. By contrast, \citet{anthropic_skills} propose \emph{skills}, which externalize execution experience in natural language form and make the underlying procedure visible to the model. This white-box property has recently attracted growing interest and shown promise across domains~\citep{li2026skillsbench,xia2026skillrl}.

To compare tools and skills in the web domain, we convert the Python tool functions of \skillweaver~into skill-like semantic knowledge via LLM prompting (see Appendix~\ref{appendix:tool_to_skill} for details and examples), and provide these skills to the agent as high-level guidance instead of executable tool calls. 

Figure~\ref{fig:tool_vs_skill} reports the results. For GPT-5, skills outperform both tools and the no-tool baseline, even though the original tools, which were created by GPT-4o, lead to slight performance drops relative to the no-tool baseline (as also discussed in \cref{subsec:tool_cannot_always_gains}). After translating the same tool knowledge into semantic skill descriptions, however, GPT-5 is able to exploit it more effectively.
% Notably, this is consistent with our earlier finding that the tools of \skillweaver~were synthesized by a weaker model (GPT-4o), and therefore do not consistently benefit a stronger user model such as GPT-5.
We interpret this advantage of skills as stemming from their transparency and flexibility. Because skills expose the procedural knowledge in natural-language form, the agent can inspect, partially follow, revise, or ignore them during planning and execution. This provides substantially more fault tolerance than hard-coded tools.
% , whose internal procedures remain hidden.
However, this pattern does not hold for GPT-5-nano. For weaker models, the burden of reasoning over and selectively applying skills may outweigh their benefits; in such cases, directly executable tools can still be more effective than semantic guidance alone.
% once a tool call fails or behaves suboptimally, the agent has little visibility into how to recover beyond retrying or changing parameters

Based on the observations in \cref{subsec:tool_cannot_always_gains} and \cref{subsec:tool_as_capacity_distil}, together with the high cost of constructing tools, we offer the following empirical suggestion for future developers. When budget constraints make it impractical to synthesize tools with strong models, a useful compromise is to translate weak-model-generated tools into semantic skills. For backbone models with sufficient reasoning capacity, such skill-form knowledge can be more effective than directly using the corresponding black-box tools, as it may better expose useful procedural knowledge while mitigating some of the limitations of weak-model-synthesized tools. 
% \textcolor{red}{not sure last paragraph, maybe just conclude skill is an alternative option for tool}
% 

\subsection{Vision Remains Beneficial, While Tools Reduce Dependence on It}
\label{subsec:vision_benefits}

\begin{table}[t!]
\centering
\setlength{\belowcaptionskip}{1.5pt}
\caption{Ablation of visual grounding (i.e., webpage screenshots) on \webarena.}
\label{tab:vision_ablation}

\normalsize
\setlength{\tabcolsep}{6pt}
\renewcommand{\arraystretch}{1.08}

\begin{tabular}{lllll}
\toprule
& \multicolumn{2}{c}{\textbf{\textit{\large CMS}}}
& \multicolumn{2}{c}{\textbf{\textit{\large Reddit}}} \\
\cmidrule(lr){2-3} \cmidrule(lr){4-5}
& \multicolumn{1}{c}{\textbf{w/ vision}} 
& \multicolumn{1}{c}{\textbf{w/o vision}}
& \multicolumn{1}{c}{\textbf{w/ vision}} 
& \multicolumn{1}{c}{\textbf{w/o vision}} \\
\midrule

\rowcolor{LightGrayRow}
\textbf{\walt} & & & & \\
\hspace{0.8em}w/o tools & 52.2 & 46.2~{\small (\MydownArrow{6.0})} & 47.4 & 37.7~{\small (\MydownArrow{9.7})} \\
\hspace{0.8em}w/ tools  & 57.1 & 50.0~{\small (\MydownArrow{7.1})} & 39.5 & 36.8~{\small (\MydownArrow{2.7})} \\

\midrule

\rowcolor{LightGrayRow}
\textbf{\skillweaver} & & & & \\
\hspace{0.8em}w/o tools & 43.4 & 37.4~{\small (\MydownArrow{6.0})} & 31.1 & 26.4~{\small (\MydownArrow{4.7})} \\
\hspace{0.8em}w/ tools  & 42.9 & 38.5~{\small (\MydownArrow{4.4})} & 30.2 & 28.3~{\small (\MydownArrow{1.9})} \\

\midrule

\rowcolor{LightGrayRow}
\textbf{\hybrid} & & & & \\
\hspace{0.8em}w/o tools & 46.7~{\small (\MyUpArrow{7.1})} & 39.6 & 42.5~{\small (\MyUpArrow{3.8})} & 38.7 \\
\hspace{0.8em}w/ tools  & 52.7~{\small (\MyUpArrow{2.2})} & 50.5 & 58.5~{\small (\MyUpArrow{1.0})} & 57.5 \\

\bottomrule
\end{tabular}
\end{table}

Existing work has shown that visual information (e.g., webpage screenshots) is beneficial for web agents. Even when textual representations such as the accessibility tree or DOM are already adequate for a target task, additional visual input can still complement the textual observation and yield significant gains~\citep{he2024webvoyager,kil2024dual,furutamultimodal}. However, when agents are equipped with tools, their reliance on direct page browsing may be reduced. This raises a natural question: does visual information still provide substantial benefits in the presence of tools?

% For web task, most modern web agent system adopt either web page accessibility tree or DOM, along with the page screenshot as supplementary information, for example, most frmaeowkrs on \webarena~uses ally tree or dom (see Table~\ref{tab:webarena_agent_compare_compact}). since page screenshot is optional for the tasks in \webarena, it's questionable that whether the vision supplement can boost performance, especailly, when the browser agent have acces to tools?
% unlike the tasks in \visualwebarena~where the vision input is mondatory, \webarena~doesn't requre agent to have vision input, and the page screenshot is totally optional for the agent.

Since webpage screenshots are not mandatory for tasks in \webarena, and both \skillweaver~and~\walt~adopt additional visual input as a complementary modality (see Table~\ref{tab:webarena_agent_compare_compact}), we ablate vision from these two frameworks. By contrast, because \hybrid~does not originally use visual input, we inject webpage screenshots into its prompts. 
Table~\ref{tab:vision_ablation} reports the results on CMS and Reddit.
For both \walt~and~\skillweaver, removing vision consistently leads to performance drops, indicating that visual input remains beneficial even when tools are available. At the same time, in most settings, the gap between the w/ vision and w/o vision conditions is smaller with tools than without tools, suggesting that tool use can partially reduce the agent’s reliance on direct webpage observation. A similar pattern is observed for \hybrid.
% : adding visual input generally improves performance, while the improvement becomes smaller once tools are enabled

Overall, vision and tools play complementary roles: visual input remains helpful beyond textual web representations, while tools make agents more robust to the removal of vision than browser interaction alone.

% Because of the DOM parser, the score of \walt~doesn't drop that much after removing the visual observation.

\section{Conclusion and Practical Takeaways}
\label{sec:conclusion}

% ======== short version =========
% In this work, we conduct a comprehensive investigation of tool use in web agents, revisiting and extending the conclusions of prior studies. By doing so, we aim to provide a stronger empirical foundation for future research. We summarize our key findings below.

% ============long version ===========
In this work, we conduct a comprehensive investigation of tool use in web agents, revisiting and extending the conclusions of prior studies under a more controlled and systematic evaluation setting. Our results show that the utility of tools is highly conditional: while human-developed tools consistently improve performance, the benefits of LLM-synthesized tools depend strongly on the relative strength between the tool constructor and the tool user, and tool use may also introduce notable costs and trade-offs. Beyond effectiveness, our study further reveals several practical design principles regarding tool complexity, compositionality, semantic skills, and visual grounding. We hope this work provides a stronger empirical foundation for future research on tool-use web agents. 

We summarize the key takeaways of this work here to facilitate future research:

\begin{takeawaycard}

\takeawayhead{Tool Synthesis Behaves Like One-Way Capability Distillation.~(\cref{subsec:tool_cannot_always_gains} \& \cref{subsec:tool_as_capacity_distil})}
\begin{takeawayitems}
    \item Human-crafted tools consistently improve web-agent performance.
    \item LLM-synthesized tools mainly benefit backbone models that are \textbf{clearly weaker than the tool constructor}; for stronger models, their utility becomes limited, inconsistent, or even negative.
\end{takeawayitems}

\takeawayhead{Tools Are Not Inherently Better When Made More Comprehensive.~(\cref{subsec:tool_cannot_over_complex})}
\begin{takeawayitems}
    \item Overly task-specific, end-to-end comprehensive tools often suffer from poor generalizability, low utility, and increased brittleness.
    \item A better design principle is to decouple agentic reasoning from tool execution: tools should only encapsulate \textbf{deterministic} and reusable UI operations, while the agent should retain \textbf{context-sensitive} reasoning grounded in the current UI state.
\end{takeawayitems}

\takeawayhead{Tool Design Should Focus on Functional Coverage and Compositionality.~(\cref{subsec:tool_coverage})}
\begin{takeawayitems}
    \item Neither tool comprehensiveness nor tool-set size directly guarantees better performance.
    \item What matters more is functional coverage: less comprehensive tools can still be effective if they compose well and collectively cover common real-world user intentions.
\end{takeawayitems}

\takeawayhead{Tool Use May Introduce Non-Trivial Efficiency Overhead.~(\cref{subsec:tax_of_tool_efficiency})}
\begin{takeawayitems}
    \item Searching through a large tool set can substantially increase token cost.
    \item Tool use does not consistently improve action efficiency, as additional steps may be spent on tool selection, error recovery, and parameter adjustment.
\end{takeawayitems}

\takeawayhead{Skills Are a Flexible Alternative to Tools.~(\cref{subsec:skill_vs_tool})}
\begin{takeawayitems}
    \item Semantic skills provide transparency and flexibility, allowing strong models to inspect, partially follow, revise, or ignore the guidance.
    \item However, skills place greater demands on the backbone model; weaker models may benefit less from skills than from directly executable tools.
\end{takeawayitems}

\takeawayhead{Vision and Tools Play Complementary Roles.~(\cref{subsec:vision_benefits})}
\begin{takeawayitems}
    \item Visual input remains beneficial even when tools are available.
    \item Tool use makes agents more robust to missing visual grounding, but does not eliminate the value of it.
\end{takeawayitems}

\end{takeawaycard}

\section*{Ethics Statement}
Web agents have the potential to improve productivity and accessibility by assisting users with routine browser interactions and complex web tasks. Our work provides a systematic empirical analysis of tool use in web agents to support future research and development toward more reliable and transparent browser‑use AI systems. At the same time, increasingly capable web agents may pose risks if deployed without proper safeguards, particularly in contexts involving sensitive online operations. Our study is not intended to promote unrestricted automation, but rather to clarify the limitations, trade‑offs, and practical implications of tool use. We hope these insights help guide future systems to be not only more capable, but also more robust and responsible.

\section*{Reproducibility Statement}
In this work, we conduct comprehensive experiments to develop a systematic understanding of tool use in web agents and to derive practical guidance for better tool-use agent design. As a result, the reproducibility of the three frameworks studied in this paper is of central importance. To ensure the reliability of our analysis and the transparency of our experimental process, we provide detailed reproduction notes in Appendix~\ref{appendix:reproduc_details}. Specifically, we document the reproduction procedures, the necessary code modifications made to the three frameworks, and potential confounding factors that may affect our conclusions.

Meanwhile, we also provide additional technical details of these frameworks in Appendix~\ref{appendix:frameowkr_details_tool_exp} that are not thoroughly discussed in the original papers, in order to provide a more transparent and informative reference for future research.

\bibliography{colm2026_conference}
\bibliographystyle{colm2026_conference}

\newpage

\appendix
\section*{Appendices}
This supplementary material elaborates on the following aspects:

\begin{itemize}
\item Appendix \ref{appendix:reproduc_details}: Reproduction Details and Statement
\item Appendix \ref{appendix:frameowkr_details_tool_exp}: Additional Framework Details and Tool Examples
\item Appendix \ref{appendix:tool_complexity_level}: Tool Complexity Levels
\item Appendix \ref{appendix:tool_to_skill}: Converting Tool Functions to Semantic Skills
% \item Appendix \ref{appendix:more_exp_result_in_main_txt}: More Experiment Results
% \item Appendix \ref{appendix:takeways_in_main_txt}: Practical Takeaways
\end{itemize}

\section{Reproduction Details and Statement}
\label{appendix:reproduc_details}
% In this section, we share the detailed information about how we reproduce the three frameworks used in our experiments, how we adjust the source code in order to reproduce the paper-report results, and how it might affect our conclusion.

In this section, we provide detailed reproduction notes for the three frameworks used in our experiments. Since faithfully reproducing these frameworks is crucial to this work, we document how we adapted their released codebases to fit our experimental setting and, when necessary, to reproduce the performance reported in their papers. We also discuss how these modifications may affect the interpretation of our results and conclusions.

\subsection{\walt~Reproduction}

We use the official codebase of \walt.\footnote{\walt~code: \url{https://github.com/SalesforceAIResearch/WALT}} For the main experiment (i.e., Table~\ref{tab:walt_res}), we use official tools released by \cite{prabhu2025walt}.\footnote{\walt's released tools: \url{https://github.com/SalesforceAIResearch/WALT/tree/main/walt-tools}} Meanwhile, we also construct tools by ourselves (used in Table~\ref{tab:tool_created_by_stronger}) by running \walt's tool discovery workflow (following their documents).

% To reproduce the results reported in their paper, we made the following necessary adjustments to the official codebase, including issues related to environment setup, tool registration, and benchmark task configurations. We also corrected implementation problems affecting website login (cookie validation issues), hard-coded URLs inside tools, and missing visual assets (for \visualwebarena).
To reproduce the results reported in their paper, we made several necessary adjustments to the official codebase, including fixes related to environment setup, tool registration, and benchmark-specific task configurations. We also corrected implementation issues affecting website login (e.g., cookie validation), hard-coded URLs inside tools, and missing visual assets in \visualwebarena.
% \begin{itemize}
%     \item Fix the cookie validation issue in both \webarena~and~\visualwebarena, since a considerable number of tasks require the agent to log into the websites automatically.
%     \item Fix the hard-coded URLs in \walt's tools, which were tied to the original authors' server hostnames.
%     \item Fix the issue preventing the framework from successfully registering tools.
%     \item Fix the missing image issue in \visualwebarena.
%     \item Fix the configuration issue in the test tasks.
% \end{itemize}

We use the same setup (GPT-5 as visual planner and GPT-5-mini as browser agent) and successfully reproduce the similar performance as reported by \cite{prabhu2025walt}. When running our experiments, we use the same model for both the planner agent and the browser agent to ensure consistent model scaling (e.g., Table~\ref{tab:walt_res}).

\subsection{\skillweaver~Reproduction}
We use the official codebase of \skillweaver.\footnote{\skillweaver~code: \url{https://github.com/OSU-NLP-Group/SkillWeaver}} For the main experiment (i.e., Table~\ref{tab:skillweaver_res}), we use official tools released by \cite{zheng2025skillweaver}.\footnote{\skillweaver's released tools: \url{https://github.com/OSU-NLP-Group/SkillWeaver/tree/main/skillnet}} While for Table~\ref{tab:tool_created_by_stronger}, we construct the tools by ourselves by utilizing \skillweaver's tool exploration pipeline.

As \cite{zheng2025skillweaver} only conduct experiments on \webarena, to enable \skillweaver~to run on \visualwebarena, we made necessary modifications to the original codebase.\footnote{The codebase of \skillweaver~already partially supports \visualwebarena: most of the classes and pipeline components for \visualwebarena have already been implemented in their repository, despite the fact that they do not report results on \visualwebarena in their paper.}

We follow the same setup as \citet{zheng2025skillweaver}, using GPT-4o as the backbone model, and reproduce results similar to those reported in their paper.

\subsection{\hybrid~Reproduction}
We use the official codebase of \hybrid.\footnote{\hybrid~code: \url{https://github.com/yueqis/API-Based-Agent}} For the main experiment (i.e., Table~\ref{tab:hybrid_res}), we use official API documents released by \cite{song2025beyond}.\footnote{\hybrid's released API documents: \url{https://github.com/yueqis/API-Based-Agent/tree/main/workspace}} 

Since the original \hybrid~framework only takes textual web observations as input (as shown in Table~\ref{tab:webarena_agent_compare_compact}), whereas \visualwebarena~requires visual inputs, we modified the codebase to support experiments on \visualwebarena by inserting task images into the agent prompts and truncating the web-state observation when necessary to control token cost. Although adapting a text-only framework to a vision-based benchmark may introduce implementation-related risks, we took care to keep the modifications minimal and avoid introducing additional confounding factors. We therefore believe that the resulting performance distribution of \hybrid~on \visualwebarena is sufficiently reliable for our analysis, and that our conclusions remain valid.

\subsection{Other Implementation Details}
\label{appendix:implementation_details}

We use Azure API services for our experiments.\footnote{\url{https://azure.microsoft.com/}} The deployment IDs of the models used are listed below: GPT-5: \texttt{gpt-5\_2025-08-07}; GPT-5-mini: \texttt{gpt-5-mini\_2025-08-07}; GPT-5-nano: \texttt{gpt-5-nano\_2025-08-07}; Grok-4.1-R: \texttt{grok-4-1-fast-reasoning}; Mistral-L-3.1: \texttt{Mistral-Large-3\_1}~\citep{GPT5,mistral3_2025,grok41_2025}. To ensure reproducibility, we set the temperature of all agents to 0 and keep the default reasoning effort at medium. All other generation parameters are kept at the default values specified in each framework's released repository.

\section{Additional Framework Details and Tool Examples}
\label{appendix:frameowkr_details_tool_exp}

% In this section, we share more details about the three frameworks used in our experiment for reliable reference and solid transparent reference for the future research.
In this section, we provide additional details (technical level) about the three frameworks used in our experiments. Although each framework has its own paper to introduce the methodology, many implementation and reproduction-related details are not fully documented in the original publications. We therefore share our observations and reproduction notes to provide greater transparency for our experimental setup and results, and to offer a more open and informative reference for future research. We hope these information can contribute to a more transparent and fair research community.

\subsection{Framework Details Not Fully Documented in the Original Papers}
Consistent with Table~\ref{tab:webarena_agent_compare_compact}, we provide additional details about the three frameworks used in our experiments. These details supplement the high-level comparison in Table~\ref{tab:webarena_agent_compare_compact} and offer a more complete understanding of each framework.

\begin{itemize}
    \item \textbf{Web observation}: \hybrid~uses only textual web observations, constructed by directly concatenating the flattened DOM and accessibility tree with simple truncation, which often results in lengthy and redundant inputs. In contrast, both \skillweaver~and~\walt~use a single structured representation—either a simplified accessibility tree or a simplified DOM—together with visual input from the current webpage screenshot. Based on this comparison, we suggest that future web agents adopt more concise textual observations, with necessary pruning of web-structure information, while preserving visual input for the model.
    \item \textbf{Tool source}: In Table~\ref{tab:webarena_agent_compare_compact}, we report that the tools used in \skillweaver~and~\walt~are constructed by GPT-4o and GPT-5-mini, respectively. However, the actual tool-construction pipeline in \skillweaver~involves both GPT-4o and o3-mini: GPT-4o is used to generate the trajectories encapsulated in the tools, while o3-mini is used in later stages of the construction process. Since the core task procedures encoded in the tools are derived from GPT-4o-generated trajectories, we keep GPT-4o as the reported tool source in the table. For \walt, neither the paper nor the released repository explicitly specifies which model was used to construct the tools. We therefore use the default \texttt{llm\_name} argument in their tool discovery workflow as our best available evidence. 
    \item \textbf{UI selectors}: At runtime, \walt~uses a cascaded selector strategy to resolve action targets. It first relies on DOM/CSS selectors, then falls back to XPath-based selectors when necessary, and finally uses \texttt{elementHash} as a backup identifier tied to previously demonstrated elements. This design makes tool execution more robust to selector instability. While \skillweaver~uses accessibility-centric selectors based on semantic roles, which can be ambiguous and may resolve to multiple elements. To alleviate this issue, the \skillweaver~forces the agent to use the full path of the accessibility node (in the prompt). Nevertheless, we observe that a substantial number of selectors still suffer from ambiguity in practice.
\end{itemize}

% ==================== recover after uppon accept =======
% \subsection{Notes on~\walt}
% Besides what has been introduced in \citep{prabhu2025walt}, after reviewing the \walt's codebase, we find that \walt~adopts a additional narrative memory mechanism that persists across tasks and is retrieved to guide future decisions. Our inspection suggests that this memory is not purely based on the agent’s own reasoning or trajectory history: it also incorporates feedback derived from task evaluation scores (calculated by \webarena's evaluation scripts), which is converted into success/failure signals and then stored as part of the memory. As a result, subsequent tasks can be influenced by evaluator-conditioned memories of prior trajectories, which may introduce unfair advantages when comparing \walt~with other frameworks that do not use such feedback-driven memory.
% % \walt~uses an additional memory techniques in their offical codebase (which doesn't mentioned in their paper). What's more, after investigating the code, we found \walt~leaked the 

% However, since our study focuses on tool use in web agents rather than solely on comparing framework performance, we keep this memory mechanism enabled in \walt, as it does not affect the scope of our investigation.

% \subsection{Notes on \skillweaver}

\subsection{Tool Examples}
\begin{figure}[ht!]
\setlength{\abovecaptionskip}{-5pt}
\begin{center}
\centering
\includegraphics[width=\textwidth]{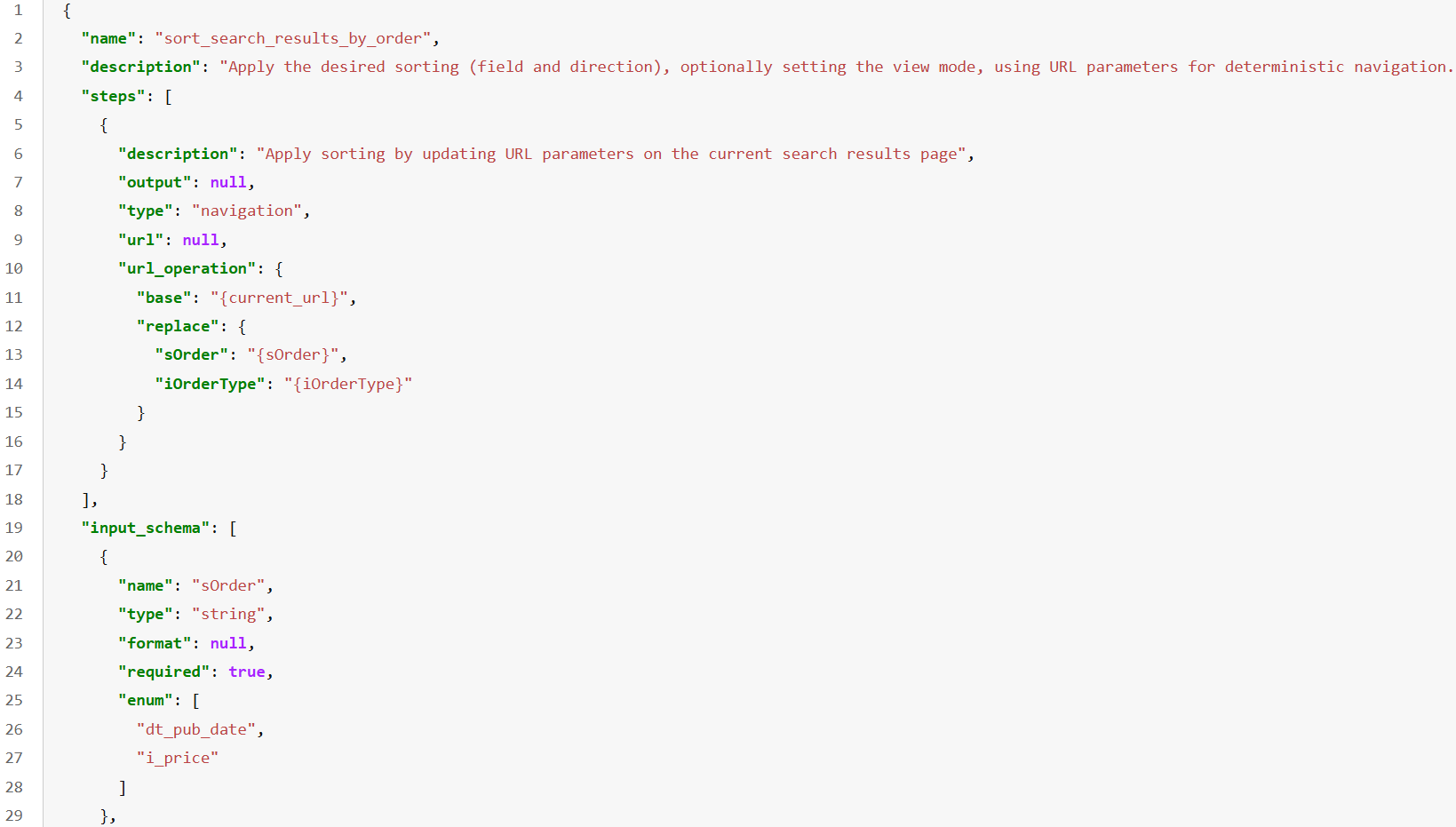}
\end{center}

\caption{Tool example of \protect\waltcolor.}
\label{fig:walt_tool_exp}
\end{figure}

\begin{figure}[ht!]
\setlength{\abovecaptionskip}{-5pt}
\begin{center}
\centering
\includegraphics[width=0.6\textwidth]{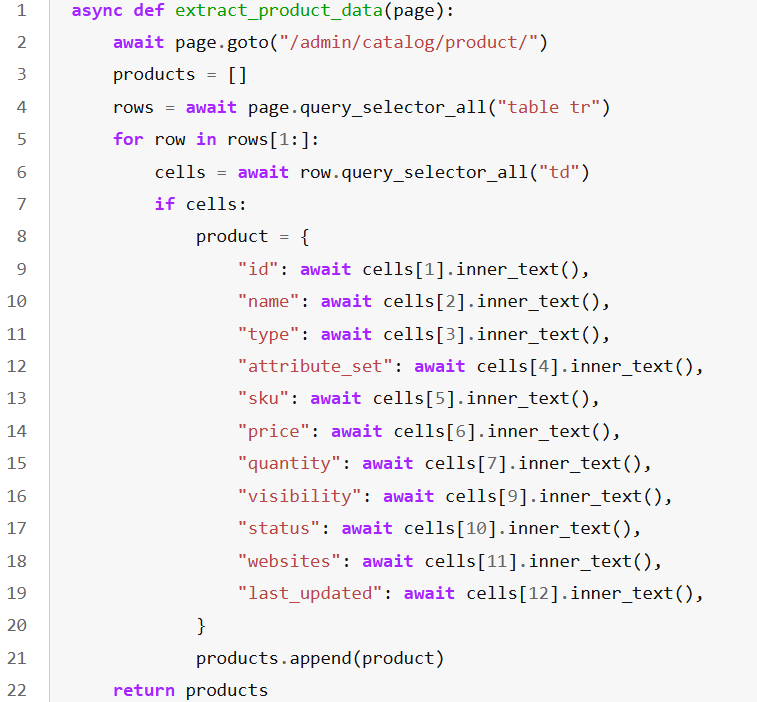}
\end{center}

\caption{Tool example of \protect\skillweavercolor.}
\label{fig:skillweaver_tool_exp}
\end{figure}

\begin{figure}[ht!]
\setlength{\abovecaptionskip}{-5pt}
\begin{center}
\centering
\includegraphics[width=0.66\textwidth]{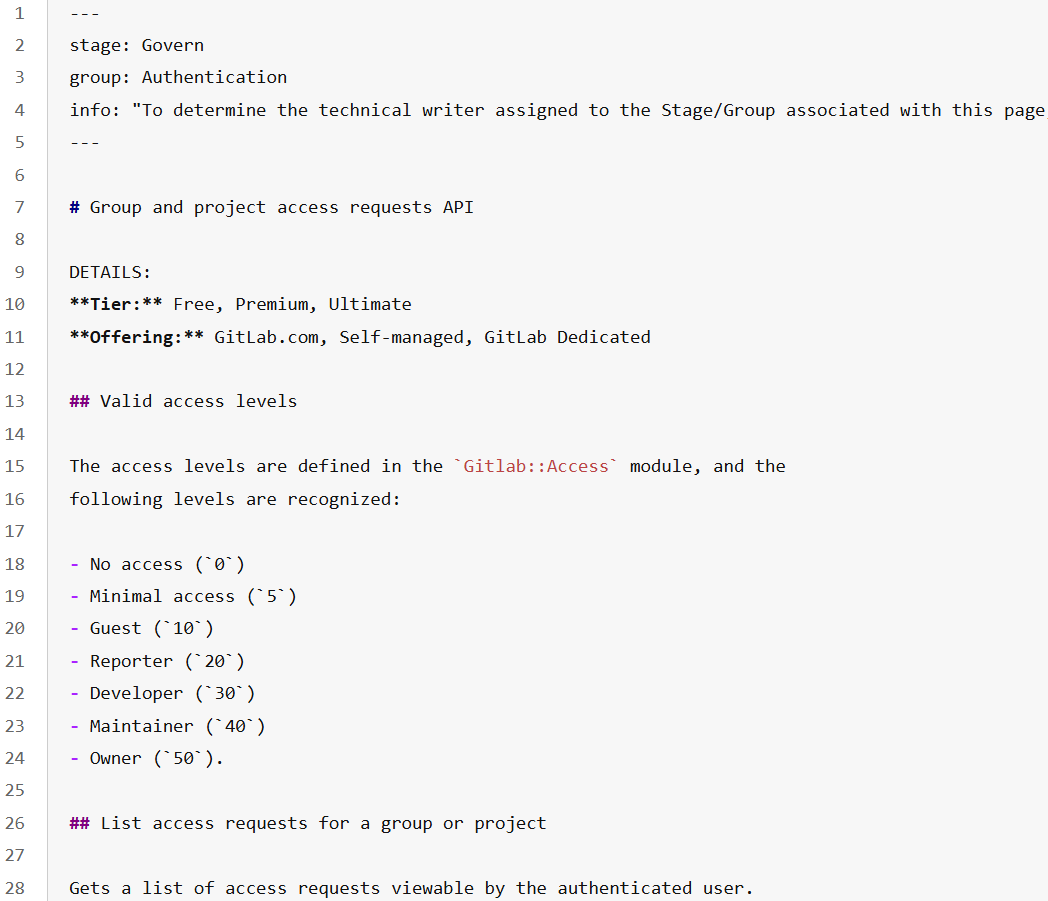}
\end{center}

\caption{Tool example of \protect\hybridcolor.}
\label{fig:hybrid_tool_exp}
\end{figure}

We select one representative tool example for each framework studied in this work to help readers better understand their design and usage in tool-use web agents, 
as shown in Figure~\ref{fig:walt_tool_exp},~\ref{fig:skillweaver_tool_exp}, and~\ref{fig:hybrid_tool_exp}.

% \begin{itemize}
%     \item \waltcolor~tool example: \url{https://github.com/SalesforceAIResearch/WALT/blob/main/walt-tools/classifieds/create_listing/create_listing.tool.json}
%     \item \skillweavercolor~tool example: \url{https://github.com/OSU-NLP-Group/SkillWeaver/blob/main/skillnet/cms/cms_kb_post_code.py#L23}
%     \item \hybridcolor~tool example: \url{https://github.com/yueqis/API-Based-Agent/blob/main/workspace/gitlab_api/access_requests.md}
% \end{itemize}

% \waltcolor~tool example:

% \skillweavercolor~tool example:

% \hybridcolor~tool example:

\section{Tool Complexity Levels}
\label{appendix:tool_complexity_level}

As shown in Figure~\ref{fig:tool_complexity_pie}, we analyze the complexity distribution of the tools used in the three frameworks. For \walt~and~\skillweaver, tool complexity is categorized by prompting GPT-4.1 (see \ref{appendix:complexity_level_prompt} for the prompt).

For \hybrid, however, the tool set is substantially larger and mostly consists of HTTP requests (e.g., \texttt{GET}, \texttt{PUT}, and \texttt{POST}). We therefore adopt a heuristic-based method to efficiently classify each API based on the API documentation provided by \citet{song2025beyond}. \ref{appendix:hybrid_tool_complexity} share more details about the heuristic-based rule.

\subsection{Prompt for Tool Complexity Classification}
\label{appendix:complexity_level_prompt}

The prompt used for tool complexity level classification is provided below:

\begin{promptbox}
\begin{lstlisting}[style=promptstyle]
You are learning how to use a website by studying reusable skills implemented as Python Playwright functions.

You will be given a Python function that represents a web automation skill:
a sequence of UI interactions that accomplishes a business-level task on the website.

Your goal is to decide the **level** of the given skill function.
The level reflects the action complexity of the skill: higher level means the task requires more steps or more complex logic.

<skill_function>
{tool_source}
</skill_function>

### Guidelines

1. Read the function docstring to understand the intended web task.
2. Analyze the function body to estimate the number of interaction steps and the presence of complex logic (e.g., loops, conditionals).
3. Classify the function into one of the following levels:

- **high**:
  - Accomplishes a complex or end-to-end web task (e.g., `search_product_and_change_price`), which are over-specific and comprehensive.
  - Involves many steps (typically more than 8) and/or complex control logic (loops, conditionals).

- **medium**:
  - Accomplishes a simple or focused web task (e.g., `search_product`, `change_price`), more like a sub-task of a complex operation, more generalizable.
  - Still involves multiple interaction steps (typically more than 5), but with limited complexity.

- **low**:
  - Accomplishes a very simple web action (e.g., `navigate_to_orders_page`), atomic and easy to generalize.
  - Involves only a few steps (typically fewer than 5) and no complex logic.

### Tips
- So, if a function contains complex logic (e.g., loops or conditionals) that strongly depends on specific UI states (i.e., brittle UI assertation), you can generally classify it as high-level.
- otherwise, focus on the task-level complexity and number of steps.

### Output Format

Return a JSON object with the following fields:

- `step_by_step_reasoning`: briefly explain how you determined the level based on the task and code structure.
- `level`: one of "high", "medium", or "low".
\end{lstlisting}
\end{promptbox}

\subsection{Heuristic Rules for Complexity Classification of \hybrid's REST APIs}
\label{appendix:hybrid_tool_complexity}

We estimate the complexity of \hybrid's REST APIs using a lightweight heuristic rule set. 
The core idea is to adapt our action-complexity rubric for functions to API tools by considering how broad an API operation is, how many inputs or constraints it requires, and whether the documentation suggests multi-step or workflow-level behavior.

Concretely, we group APIs into three levels: 
\begin{itemize}
    \item \textbf{Low}-complexity APIs correspond to atomic operations, such as simple retrieval, deletion, or single-resource updates. 
    \item \textbf{Medium}-complexity APIs correspond to more focused but non-trivial operations, such as search, filtering, pagination, or structured create/update actions. 
    \item \textbf{High}-complexity APIs correspond to more comprehensive or end-to-end behaviors, such as bulk processing, import/export, migration, asynchronous operations, or workflow-like procedures spanning multiple coordinated steps.
\end{itemize}

Our heuristic classification mainly relies on several high-level signals from the API documentation, including the HTTP method type, the semantic scope implied by documentation keywords, the number of explicit parameters, and whether the document describes a single endpoint or a multi-step workflow. 
Intuitively, APIs that are broader in scope, require more inputs, or imply orchestration across multiple steps are assigned higher complexity. 
Based on these signals, each API is mapped to one of the three complexity levels using fixed thresholds.

\section{Converting Tool Functions to Semantic Skills}
\label{appendix:tool_to_skill}

\begin{figure}[!htbp]
% \vspace{-2.3em}
\setlength{\abovecaptionskip}{-6pt}
\begin{center}
\centering
\includegraphics[width=1.01\textwidth]{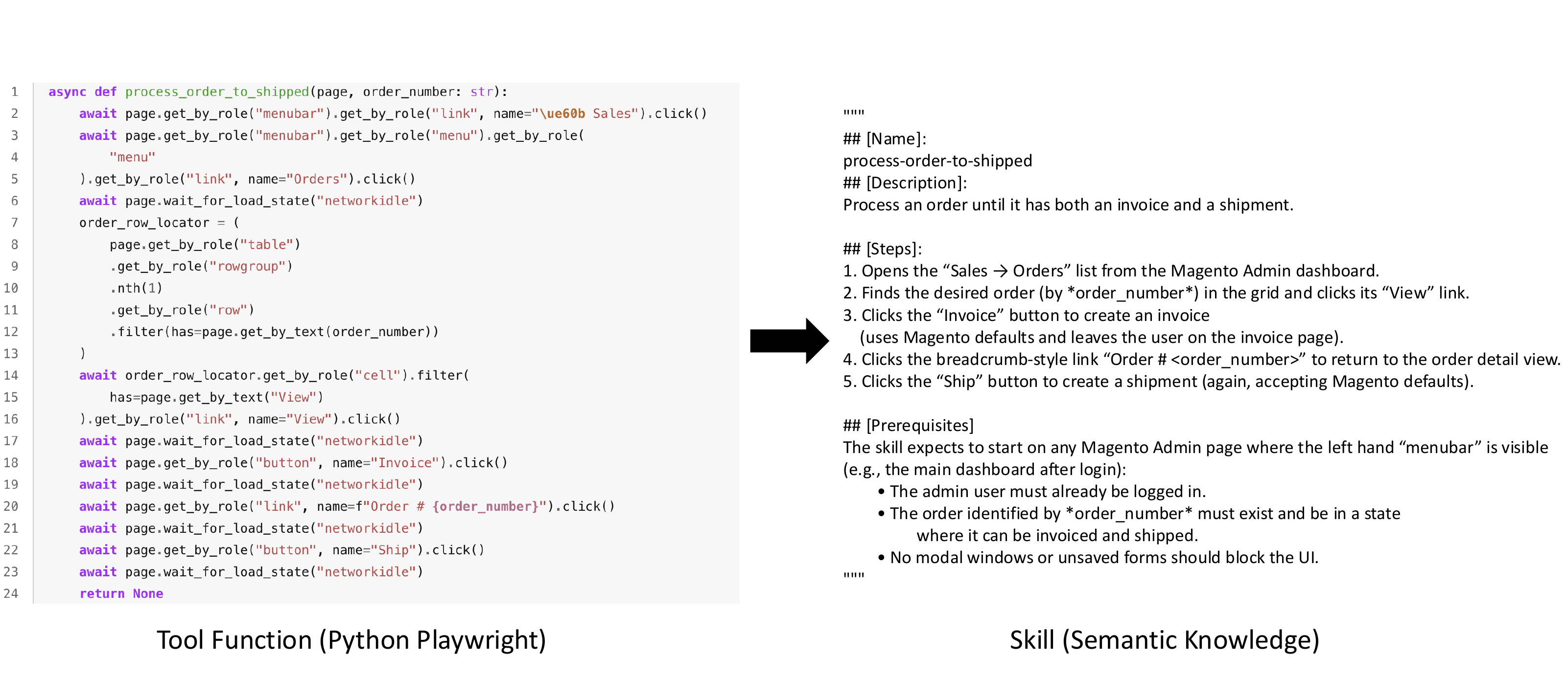}
\end{center}

\caption{Example of the Python tool function of \skillweaver~and the converted skill.
}
\label{fig:tool_to_skill}
\end{figure}

% We display an example of the tool function of \skillweaver~and it's converted skills, as shown in Figure~\ref{fig:tool_to_skill}.
We show an example tool function from \skillweaver~and its converted semantic skill, as shown in Figure~\ref{fig:tool_to_skill}.

We use GPT-4.1 to ``translate'' Python tool functions of \skillweaver~to equivalent skills (semantic knowledge), by using the following prompt: 

\begin{promptbox}
\begin{lstlisting}[style=promptstyle]
You are learning how to use a website by studying skills (Python Playwright functions).

You will be given a Python function that represents a reusable skill:
a sequence of UI interactions that accomplishes a meaningful business-level task on the website.

Your goal is to extract the **semantic knowledge** of this skill:
a high-level, natural-language description of the **procedural steps** a human would follow to accomplish the same task on the website.

<skill_function>
{function_source}
</skill_function>

### Guidelines

- The function consists of multiple Playwright actions, each interacting with a specific UI element.
- Carefully analyze the code to understand **what task is being performed** and **in what order**.
- Describe the procedure from a **human / user perspective**, not from an implementation perspective.
  - Focus on *what the user does*, not *how Playwright locates elements*.
  - Do NOT mention selectors, API calls, or code-level details (e.g., `get_by_role`, `click`, etc.).
- The extracted steps should be generalizable and task-oriented, while allowing references to meaningful UI components
  (e.g., search boxes, input fields, buttons) when they are helpful for understanding how to perform the task, but avoiding low-level implementation or layout-specific details.
  - Here's an example of good semantic knowledge: `1. Opens the "Orders" section from the admin navigation menu. 2. Locates the search input field on the orders page. 3. ...`
- If the function already contains a docstring describing the procedure:
  - Reuse it directly as the final semantic knowledge if it is accurate and complete.
- Otherwise:
  - Infer the procedural steps strictly from the code.
  - Do NOT invent steps that are not supported by the function.

### Output Format

Return a JSON object with the following fields:

- `step_by_step_reasoning`:
  A brief explanation of how you analyzed the function to understand its behavior.
  This is for transparency only and should be concise.

- `semantic_knowledge`:
  A clear, ordered list of procedural steps (in natural language),
  describing how a human would accomplish the same task on the website.
  This will serve as reusable semantic knowledge for future agents or users.
\end{lstlisting}
\end{promptbox}

\end{document}